\documentclass[letterpaper, 10 pt, conference]{ieeeconf}  
\IEEEoverridecommandlockouts                              
\title{\LARGE \bf
Explicit Contact Optimization in \\ 
Whole-Body Contact-Rich Manipulation \\
}

\author{
    Victor Levé$^{1}$ $^{2}$, João Moura$^{1}$, Namiko Saito$^{1}$, Steve Tonneau$^{1}$ and Sethu Vijayakumar$^{1}$ %
    \thanks{$^{1}$School of Informatics, The University of Edinburgh, UK}%
    \thanks{$^{2}$First author contact email: {\footnotesize s2522875@ed.ac.uk}}%
    \thanks{This work is supported by the JST Moonshot R\&D (Grant No. JPMJMS2031), the Kawada Robotics Corporation and The Alan Turing Institute.}%
}
\DeclareTextSymbolDefault{\dh}{T1}
\usepackage{hyperref}
\usepackage{graphicx}
\usepackage{amsmath}
\usepackage{tikz}
\usetikzlibrary{shapes.geometric, arrows, fit, backgrounds}
\usepackage{pgfplots}
\usepgfplotslibrary{fillbetween}
\usepackage{subcaption}
\usepackage{lastpage}
\usepackage{multirow}
\pgfplotsset{compat=1.18}

\begin{document}
\bstctlcite{IEEEexample:BSTcontrol}

\maketitle
\thispagestyle{empty}
\pagestyle{empty}

\begin{abstract}
Humans can exploit contacts anywhere on their body surface to manipulate large and heavy items, objects normally out of reach or multiple objects at once. 
However, such manipulation through contacts using the whole surface of the body remains extremely challenging to achieve on robots. 
This can be labelled as Whole-Body Contact-Rich Manipulation (WBCRM) problem.
In addition to the high-dimensionality of the Contact-Rich Manipulation problem due to the combinatorics of contact modes, admitting contact creation anywhere on the body surface adds complexity, which hinders planning of manipulation within a reasonable time.
We address this computational problem by formulating the contact and motion planning of planar WBCRM as hierarchical continuous optimization problems.
To enable this formulation, we propose a novel continuous explicit representation of the robot surface, that we believe to be foundational for future research using continuous optimization for WBCRM.
Our results demonstrate a significant improvement of convergence, planning time and feasibility -- with, on the average, 99\% less iterations and 96\% reduction in time to find a solution over considered scenarios, without recourse to prone-to-failure trajectory refinement steps. See more in our video: {\footnotesize \url{https://youtu.be/AfnDWBqJzsY}}.
\end{abstract}


\section{INTRODUCTION}

Recent advances in humanoid robotics research have showcased excellent locomotion skills. However, when it comes to manipulation skills, they remain limited to prehensile manipulation using a single pre-specified contact location on the end-effectors. Meanwhile, humans are capable of achieving complex manipulations using their whole body, e.g. opening a door using elbows or legs while holding a large box with their arms. These kind of manipulations through contacts utilize different contact modalities and exploit the whole surface of the body to contact with objects. We will refer to such manipulations as Whole-Body Contact-Rich Manipulation (WBCRM).

\begin{figure}[t]
    \centering
    \begin{subfigure}{0.15\textwidth} \includegraphics[width=1\linewidth]{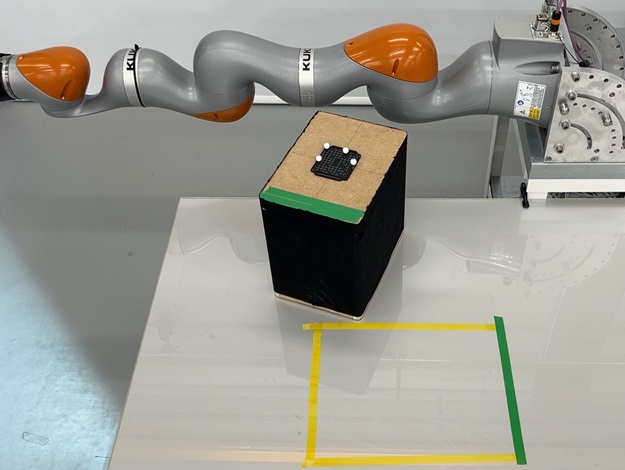} \end{subfigure} 
    \begin{subfigure}{0.15\textwidth} \includegraphics[width=1\linewidth]{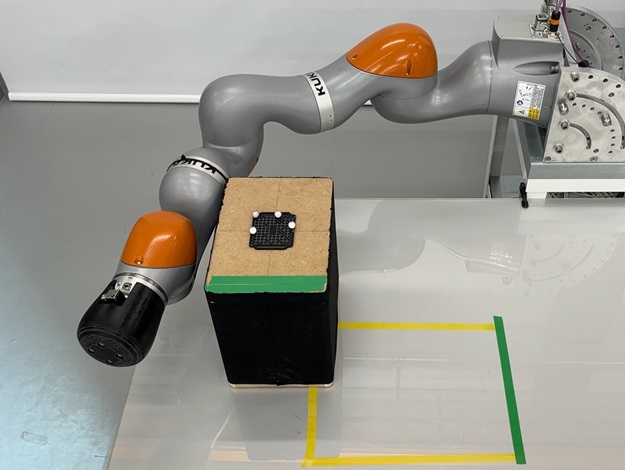} \end{subfigure} 
    \begin{subfigure}{0.15\textwidth} \includegraphics[width=1\linewidth]{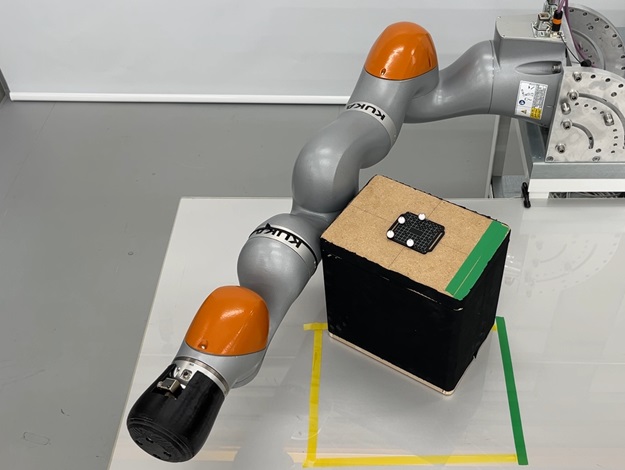} \end{subfigure} 

    \centering
    \begin{subfigure}{0.15\textwidth} \includegraphics[width=1\linewidth]{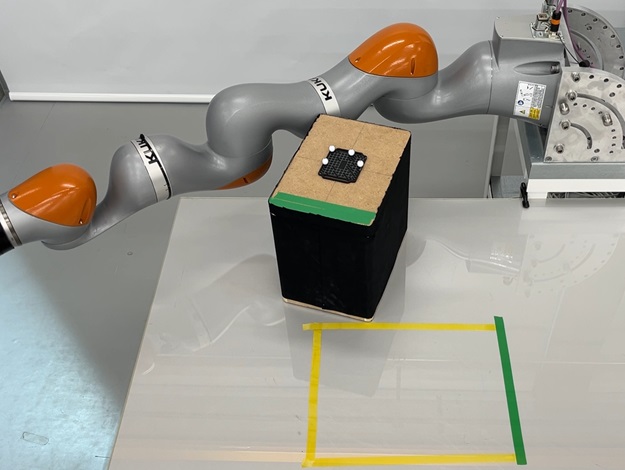} \end{subfigure} 
    \begin{subfigure}{0.15\textwidth} \includegraphics[width=1\linewidth]{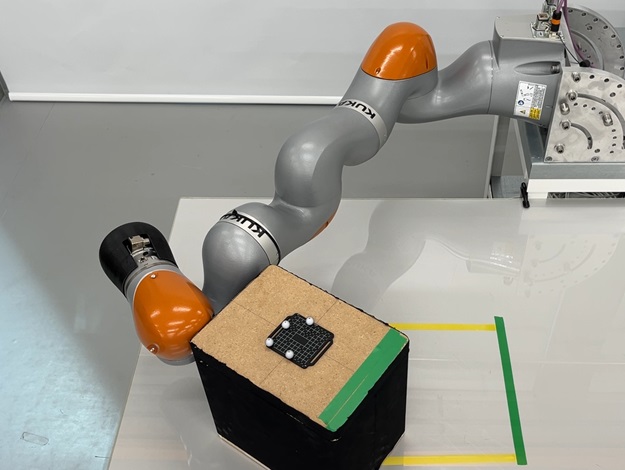} \end{subfigure} 
    \begin{subfigure}{0.15\textwidth} \includegraphics[width=1\linewidth]{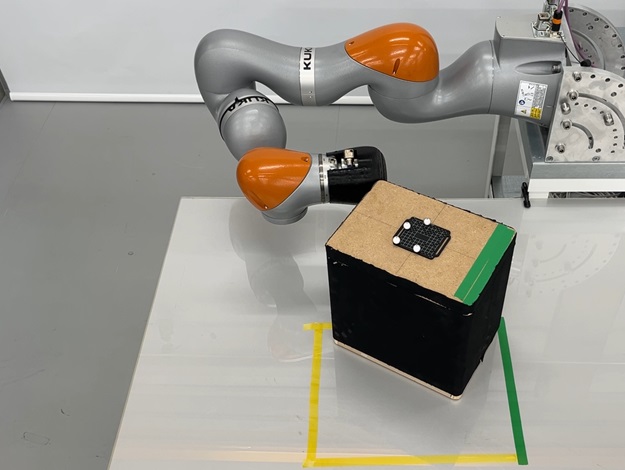} \end{subfigure} 

    \centering
    \begin{subfigure}{0.15\textwidth} \includegraphics[width=1\linewidth]{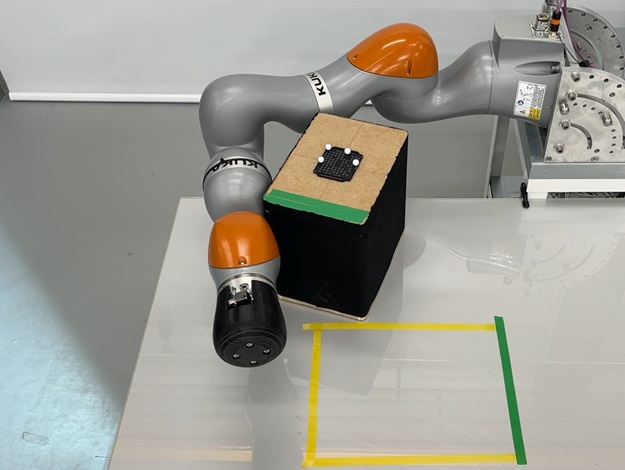} \end{subfigure} 
    \begin{subfigure}{0.15\textwidth} \includegraphics[width=1\linewidth]{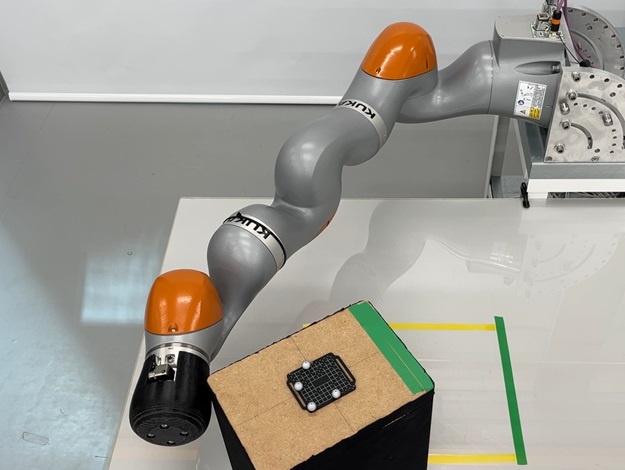} \end{subfigure} 
    \begin{subfigure}{0.15\textwidth} \includegraphics[width=1\linewidth]{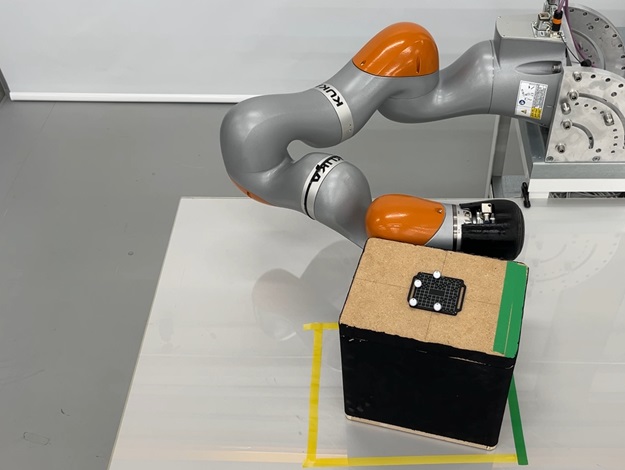} \end{subfigure} 
    
    \centering
    \captionsetup{justification=centering,margin=0.0cm}
    \caption{Snapshots of a robot performing planar Whole-Body Contact-Rich Manipulation of a box in a real setup.}
    \label{fig:exp4_figure}
\end{figure}

Enhancing the WBCRM capability of humanoid robots is essential to improve their efficiency and versatility in order to deal autonomously with unexpected situations without human support.
Such skills allow to handle large or heavy objects \cite{10417696} \cite{7139995}, under-actuated objects \cite{pang2023globalplanningcontactrichmanipulation} or even multiple objects at once \cite{xie2019improvisationphysicalunderstandingusing}.
We focus on the specific problem of contact and motion planning of planar WBCRM with re-orientation, using a planar robot, as pictured in Fig.\ref{fig:exp4_figure}.
The re-orientation ensures enough complexity in the manipulation task to make the use of whole-body contact relevant.

The combinatorial explosion due to the contact modes -- e.g. sticking, sliding, breaking -- is a well-known issue of Contact-Rich Manipulation. 
The additional dimension of the continuous surface used to make contact in WBCRM further exacerbates this problem.
This makes computationally expensive methods, such as mode enumeration \cite{cheng2022contactmodeguidedmotion} or contact location sampling \cite{nakatsuru2023implicitcontactrichmanipulationplanning}, unsuitable for scaling to whole-body contact location optimisation.

\tikzstyle{bloc3} = [rectangle, rounded corners, minimum width=3.5cm, minimum height=2.5cm, text centered, fill=white, align=center]
\tikzstyle{inout} = [anchor=south, align=center, node distance=1.5cm]

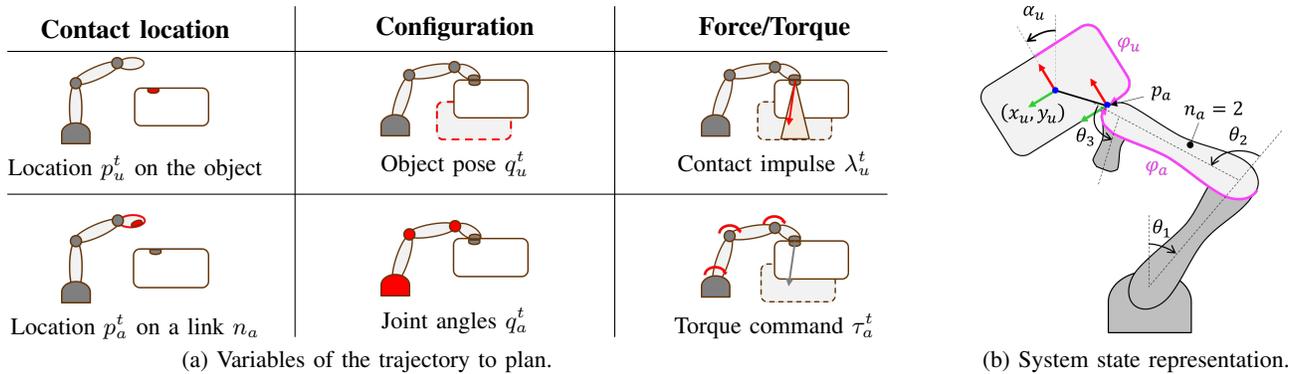
\begin{figure*}[t]
    \begin{subfigure}[t]{0.55\textwidth}
        \centering 
        \begin{tikzpicture}[node distance=4.25cm]
            \node (cntloc_u) [bloc3] {
                \includegraphics[width=.2\textwidth]{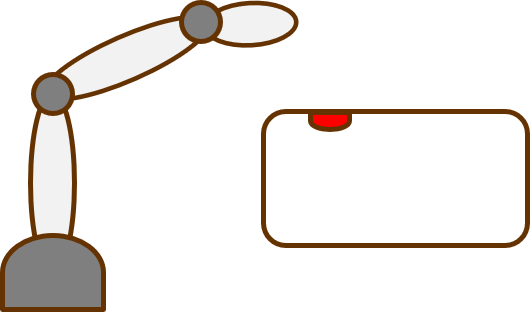} \\
                \small{Location} $p_u^t$ on \small{the object} 
                };
            \node (cntloc_a) [bloc3, below of=cntloc_u, node distance=2.1cm] 
                {
                \includegraphics[width=.2\textwidth]{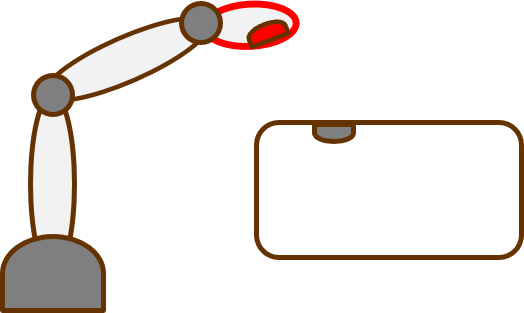} \\
                \small{Location} $p_a^t$ \small{on a link} $n_a$ 
                };
            \node (qstate_u) [bloc3, right of=cntloc_u] {
                \includegraphics[width=.2\textwidth]{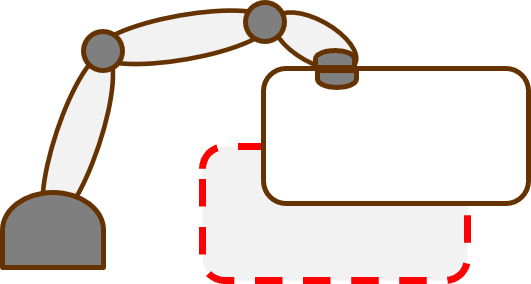} \\ 
                \small{Object pose} $q_u^t$ 
                };
            \node (qstate_a) [bloc3, below of=qstate_u, node distance=2.1cm] 
                {
                \includegraphics[width=.2\textwidth]{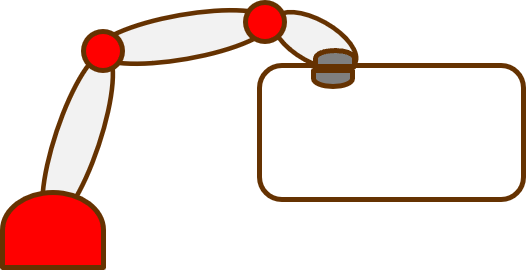} \\ 
                \small{Joint angles} $q_a^t$ 
                };
            \node (ucontr_u) [bloc3, right of=qstate_u] {
                \includegraphics[width=.2\textwidth]{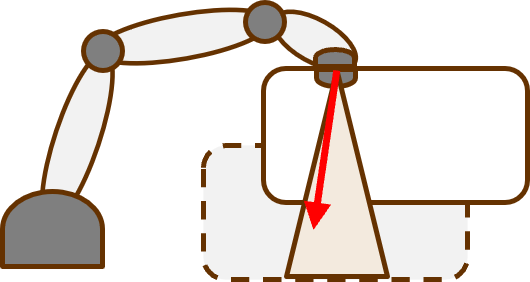} \\ 
                \small{Contact impulse} $\lambda_u^t$ 
                };
            \node (ucontr_a) [bloc3, below of=ucontr_u, node distance=2.1cm] 
                {
                \includegraphics[width=.2\textwidth]{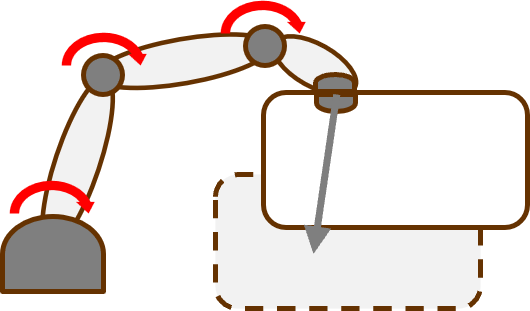} \\ 
                \small{Torque command} $\tau_a^t$ 
                };
    
            \node (col1) [inout, above of=cntloc_u, node distance=1.2cm]  {\textbf{Contact location}};
            \node (col2) [inout, above of=qstate_u, node distance=1.2cm]  {\textbf{Configuration}};
            \node (col3) [inout, above of=ucontr_u, node distance=1.2cm]  {\textbf{Force/Torque}};

            \draw ( 2.125, 1.5) -- (2.125,-2.9);
            \draw ( 6.375, 1.5) -- (6.375,-2.9);

            \draw (-1.7, 0.9) -- (10, 0.9);
            \draw (-1.7,-1.0) -- (10,-1.0);
        \end{tikzpicture}
        \centering
        
        \vspace*{-5mm}
        \captionsetup{justification=centering,margin=0.0cm}
        \caption{Variables of the trajectory to plan.}
        \label{fig:trajectory}
    \end{subfigure}%
    \hfill{}
    \begin{subfigure}[t]{0.3\textwidth}
        \centering
            \centering
            \raisebox{5mm}{\includegraphics[width=0.75\linewidth]{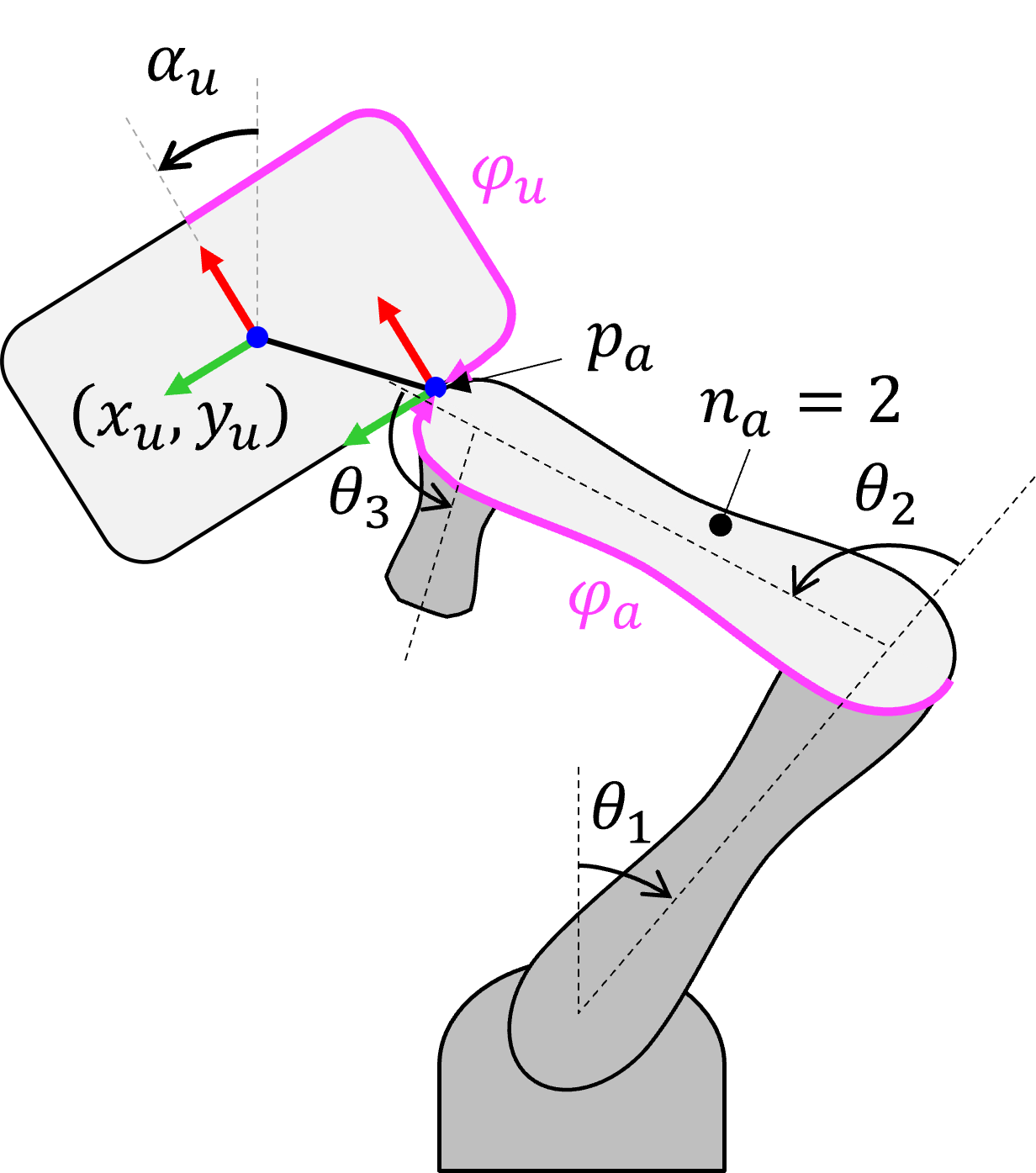}}
        \centering
        \captionsetup{justification=centering,margin=0.0cm}
        \vspace*{-5mm}
        \caption{System state representation.}
        \label{fig:staterepresentation}
    \end{subfigure}%
    \caption{Problem representation of the planar Whole-Body Contact-Rich Manipulation.}
\end{figure*}

We propose a novel continuous representation of the non-convex surface of the robot to contact with object, using Gaussian functions. 
This representation enables us to formulate a continuous optimization of the explicit contact location on the object and the robot, using any link of the robot, as well as the object pushing trajectory as a Trajectory Optimization (TO) that includes sliding on the robot surface. 
We show that, for a considered set of planar WBCRM scenarios with re-orientation, our approach reduces the average planning computation time from about 48min to 1min30s -- generating an order of magnitude improvement compared to the state-of-the-art.
It also ensures that the planned trajectory already satisfies the model constraints, which maximizes the feasibility of the plan.
Our work is demonstrated in planar settings at the moment and paves the way for future research to use continuous optimization of contact location on the whole surface of the robot to solve more complex WBCRM problems.


\section{RELATED WORK}

Contact-Rich Manipulation is well-known for its combinatorial complexity resulting from all the possible contact mode choices (sticking, rolling, sliding and breaking) at each time step, with different dynamic behaviors for each mode \cite{trinkle1997} \cite{posa2014directmethod}. The dimension added by the choice of contact location on the robot surface in WBCRM problems makes scalability a crucial prerequisite of any approach. 

Numerous works have used random sampling methods, often relying on Rapidly exploring Random Tree (RRT), to deal with the combinatorics of contact modes \cite{doi:10.1177/0278364910386985} \cite{7989390} \cite{7139995}. However, the convergence of random sampling methods to a solution is only guaranteed asymptotically, and the actual planning computation cost is too high to scale well to the dimensionality of WBCRM.

To cope with this high computation cost, several approaches use approximations (e.g. convex relaxation, smoothing) of the system kinodynamics to accelerate the computation \cite{pang2023globalplanningcontactrichmanipulation} \cite{graesdal2024tightconvexrelaxationscontactrich} \cite{howell2023trajectoryoptimizationoptimizationbaseddynamics} \cite{66470f3777564080b33178e5f9f7fae6}. Yet, these approximations degrade the quality of the solution which leads to feasibility issues when refining or executing trajectories on hardware.

Data-driven approaches with Reinforcement Learning have demonstrated success even for in-hand manipulation and WBCRM \cite{b3044c6b16944b31aa68e8850b8f118f} \cite{chen2021generalinhandobjectreorientation} \cite{zhang2023planguidedreinforcementlearningwholebody}. However, it requires a significant amount of curated data and training time, and cannot adapt well to unexpected manipulation scenarios such as pushing the object with the back of the hand or the arm, if they were not covered during the training.

Another approach has been to rely on Complementarity Constraints to formulate a TO that includes contact mode switching decision \cite{posa2014directmethod}. This approach enables long-horizon contact planning in Locomotion \cite{8283570} but the contact sliding mode, which is essential for efficient manipulation of objects, is not considered. Contact-Implicit TO combined with hydroelastic contacts has demonstrated the generation of realistic WBCRM motions \cite{9981686}, yet it can only solve contact sequences locally and its adaptability to stiff objects is unclear. In \cite{aydinoglu2022realtimemulticontactmodelpredictive} \cite{moura2022nonprehensileplanarmanipulationtrajectory} real-time Model Predictive Control has been achieved for manipulation with contact sliding mode. Yet, it is restricted to problems with low dimensionality and fails to converge in practicable time for WBCRM problems with full robot kinematic constraints. 

A recent promising approach combines discrete sampling together with TO to avoid local minima while enforcing the robot kinematic constraints and remaining computationally efficient (with planning in the order of minutes) \cite{chen2021trajectotreetrajectoryoptimizationmeets} \cite{natarajan2024longhorizonplanningcontact}. However, adapting such methods to the additional dimension of the robot contact surface stays an open problem.

We, therefore, build on this latest approach by addressing the decision of the contact location on the robot surface to the planning of planar WBCRM. Our contributions include 
1) A novel explicit representation of the contact location on the surface of the robot; 
2) The use of this representation in a contact planning formulated as a continuous optimization with explicit contact locations; 
3) The use of this representation in motion planning formulated as a two-stages TO with contact sliding on the robot surface.
\tikzstyle{bloc} = [rectangle, rounded corners, minimum width=2.0cm, minimum height=2.0cm, text centered, draw=black, fill=yellow!10, font=\footnotesize, align=center, anchor=north]
\tikzstyle{circ} = [circle, minimum width=2.0cm, minimum height=2.0cm, text centered, draw=black, fill=yellow!10, font=\footnotesize, align=center, anchor=north]
\tikzstyle{decision} = [diamond, minimum width=2cm, minimum height=1cm, text centered, draw=black, fill=lime!10]
\tikzstyle{arrow} = [thick,->,>=stealth]
\tikzstyle{inout} = [align=center, node distance=1.5cm]

\begin{figure*}[ht]
    \centering
    \begin{tikzpicture}[node distance=3.3cm]
        \node (sampling) [bloc, fill=yellow!10] {\textbf{Context Sampling} \\ 
            \includegraphics[width=0.1\linewidth]{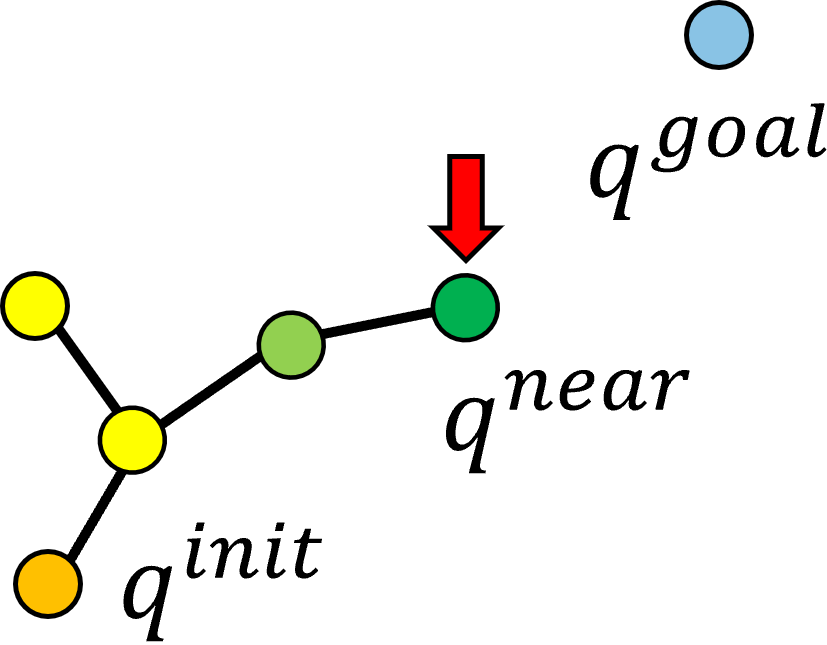}};
        \node (regrasp) [bloc, right of=sampling, fill=yellow!10] {\textbf{Contact Planning} \\\\  
            \includegraphics[width=.1\textwidth]{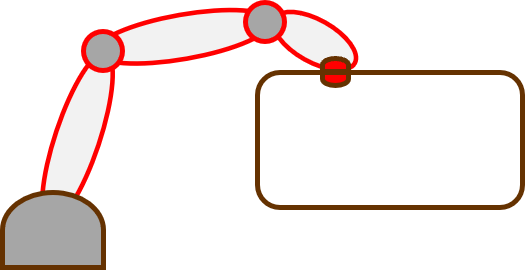}};
        \node (guidefree) [bloc, right of=regrasp, dotted, minimum height=1.6cm] {contact-free \\\\  
            \includegraphics[width=0.1\linewidth]{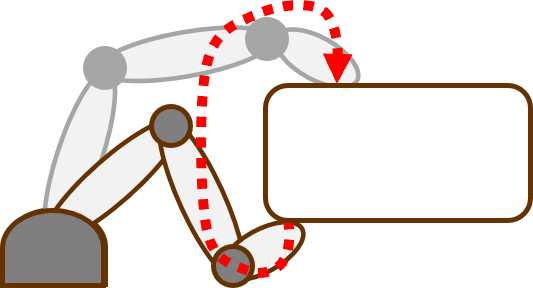}};
        \node (trackfree) [bloc, right of=guidefree, dotted, minimum height=1.6cm] {contact-free \\\\
            \includegraphics[width=0.1\linewidth]{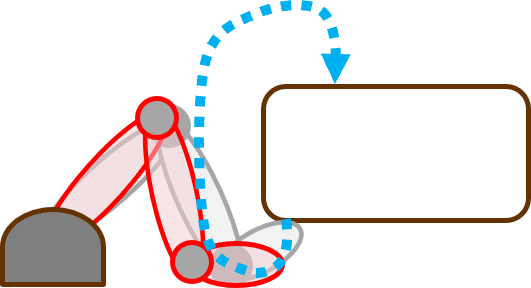}};
        \node (guide) [bloc, below of=guidefree, dotted, node distance=1.7cm, minimum height=1.6cm] {in-contact \\\\ 
            \includegraphics[width=0.1\linewidth]{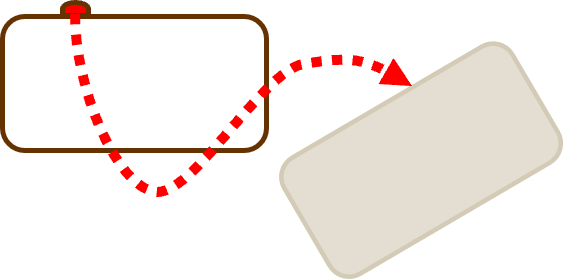}};
        \node (track) [bloc, right of=guide, dotted, minimum height=1.6cm] {in-contact \\\\ 
            \includegraphics[width=0.1\linewidth]{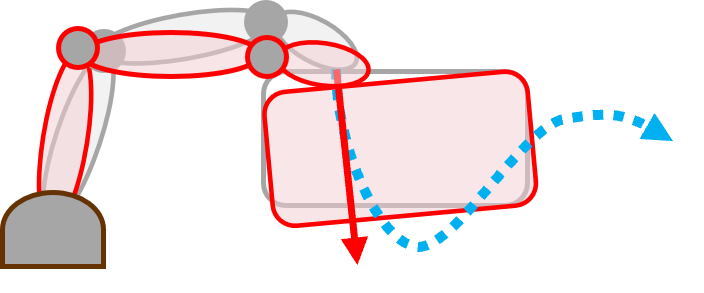}};
        
        \begin{scope}[on background layer]
            \node[fit=(guidefree) (guide), draw, inner sep=0.1cm, rounded corners, fill=yellow!10] (guides) { };
            \node[fit=(trackfree) (track), draw, inner sep=0.1cm, rounded corners, fill=yellow!10] (tracks) { };
            \node (guidestitle) [inout, above of=guides, node distance=1.8cm, anchor=south, font=\footnotesize]  {\textbf{Long-Horizon Guide}};
            \node (trackstitle) [inout, above of=tracks, node distance=1.8cm, anchor=south, font=\footnotesize]  {\textbf{Guide tracking}};

        \end{scope}

        \node (tree) [bloc, right of=tracks] {tree memory \\\\
            \includegraphics[width=.09\textwidth]{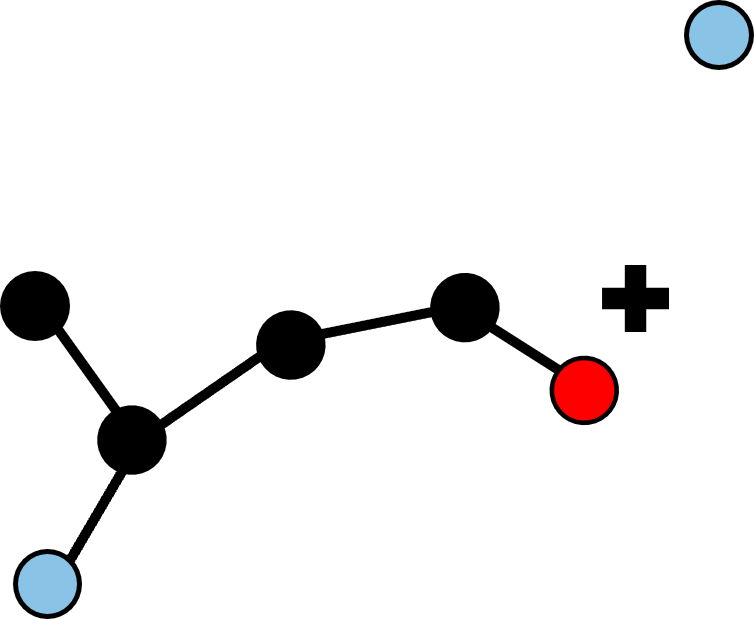} };
        
        \node (readtree) [inout, above of=sampling, font=\footnotesize, anchor=east] {read};
        \node (iterate) [inout, below of=sampling, font=\footnotesize, anchor=east] {iterate};
        \node (endplan) [inout, right of=tree, xshift=0.2cm]  {$q^t$ \\ $u^t$ \\\\ \footnotesize{end of} \\ \footnotesize{planning}};

        \draw [arrow] (sampling.west) ++(-1.0, 0.0) -- node [inout, anchor=south] {$q^{init}$ \\ $q_u^{goal}$} (sampling);
        \draw [arrow] (sampling) -- node [inout, anchor=south] {$q_u^{near}$ \\ $ n_a$} (regrasp) ;
        \draw [arrow] (regrasp) -- node [inout, anchor=south] {$\phi_u^{(0)}$ \\ $\phi_a^{(0)}$ \\ $q_a^{(0)}$} (guidefree) ;
        \draw [arrow] (guidefree.east)  -- node [inout, anchor=south] {$p_a^t$} (trackfree.west) ;
        \draw [arrow] (regrasp.east) -- ++(0.5, 0.0) |- (guide.west) ;
        \draw [arrow] (guide.east) -- node [inout, anchor=south] {$q_u^t$ \\ $u_u^t$} (track.west) ;
        \draw [arrow] (tracks.south) -- ++(0.0, -0.4) node [anchor=west, align=center, font=\footnotesize] {end of tracking} -| (sampling);
        \draw [arrow] (tree.east) -- ++(1.0, 0.0);
        
        \draw [arrow]  (tree.north) -- ++(0.0, 1.5) -| (sampling.north);
        \draw [arrow] (tracks.east)  -- node [inout] {$q^{(k+1)}$ \\ $u^{(k)}$ \\\\ \footnotesize{write} \\ \ } (tree.west) ;
        
    \end{tikzpicture}
    \centering
    \captionsetup{justification=centering, margin=2.0cm}
    \caption{Pipeline of the proposed method}
    \label{pipeline}
\end{figure*}
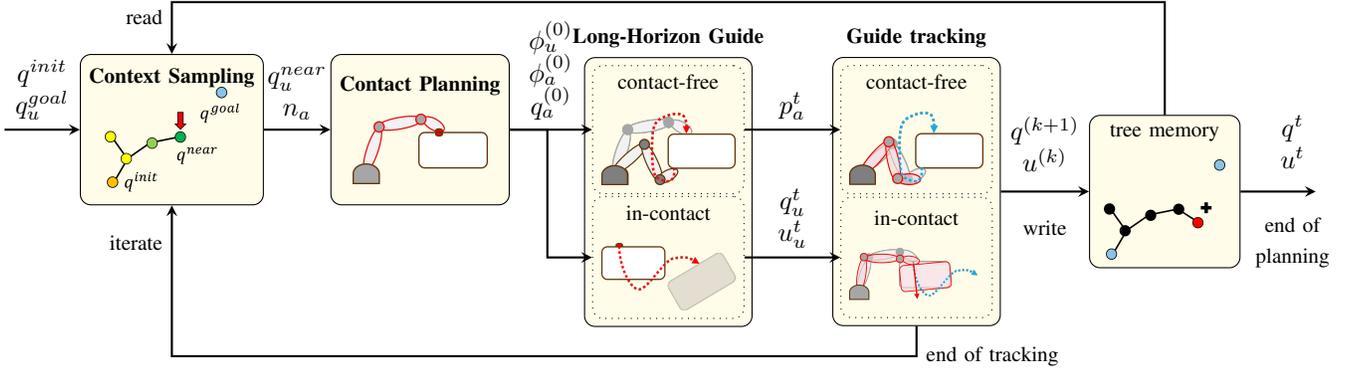
\section{PROBLEM STATEMENT}
This work addresses the problem of planar pushing of an object with re-orientation using the whole surface of a robot moving in that same plane (planar WBCRM).
This involves moving the object from an initial state $q_u^{init}$ to a final state $q_u^{goal}$ using the robot in initial configuration $q_a^{init}$, while respecting the kinematic constraints. The subscript ${\circ}_a$ denotes variables related to the actuated part of the system (the robot) while the subscript ${\circ}_u$ are used for the under-actuated part of the system (the object). 

The Fig.\ref{fig:trajectory} outlines the decision variables of this manipulation trajectory, which are 
1) the contact location $p_u^t \in {\rm I\!R}^2$ on the object; 
2) the contact location $p_a^t \in {\rm I\!R}^2$ on the link number $n_a \in [1, ..., N_a]$ of the robot -- $N_a$ being the number of actuated joints; 
3) the joints configuration $q_a^t \in {\rm I\!R}^{N_a}$; 
4) the object state $q_u^t \in {\rm I\!R}^3$ (pose and orientation); 
5) the contact impulse $\lambda_u^t \in {\rm I\!R}^2$ applied on the object; 
6) the robot joints torque $\tau_a^t \in {\rm I\!R}^{N_a}$. 
The superscript ${\circ}^t$ indicates a trajectory ${\circ}^t=(\circ^{(0)}, ..., \circ^{(N-1)})$ of length $N$.

We assume that the object shape is convex, to simplify the collision avoidance (see \ref{sub:collisionavoidance}), and the robot shape is non-convex, as it is generally the case for hardware. The object and robot motions are constrained to the plane $(e_x, e_y)$. The object state becomes $q_u^t = (x_u^t, y_u^t, \alpha_u^t)^\top$ with $(x_u^t, y_u^t)$ being the position coordinates and $\alpha_u$ the orientation. We consider the robot KUKA IIWA 14 with only the $2^{nd}, 4^{th}$ and $6^{th}$ joints axes movable ($N_a = 3$) and aligned along the axis $e_z$ orthogonal to the plane. The contact location lies on the 2D outline of each body surface and we assume the robot and the object to have only one point of contact at a time. We assume the dynamics of the system to be quasi-dynamic, i.e. velocities and acceleration are approximated to a small regularization term.


\section{METHOD}

\subsection{Pipeline}

The pipeline of the proposed method combines concepts from RRT and TO. Similar to RRT, we define a tree whose nodes encode the system state $q=(q_u^\top, q_a^\top)^\top$ and that grows according to the pipeline described in Fig.\ref{pipeline}. Given an initial robot and object states and a goal object state: 
1) The \textit{Context Sampling} randomly samples key discrete variables, for e.g., which link to make contacts with the object; 
2) The \textit{Contact Planning} decides the contact location between the object and the robot as well as the robot joint configuration to reach that contact; 
3) The \textit{Long-Horizon Guide} outputs both a contact-free trajectory of the robot contact point to reach the object contact location, and a free pushing (without the robot kinematic constraints) trajectory of the object; 
4) The \textit{Guide Tracking} tries to follow each of the trajectories prescribed by the \textit{Long-Horizon Guide} while enforcing all the robot kinematic constraints to ensure the feasibility of the planned trajectory.
If the guide is intractable, we repeat the same process again with new random samples, until reaching the goal state.
The contact-free trajectory is computed prior to the in-contact one. Each of the modules of the pipeline are explored in detail next. 

\subsubsection{Context Sampling}
\label{sub:sampling}

The \textit{Context Sampling} narrows down the number of decision variables for the next optimization problems to keep them computationally feasible. At each iteration, it selects the discrete variables $q^{near}$ -- the node from which to extend the tree -- and $n_a$ -- which link of the robot will contact with the object -- by a weighted random selection that promotes more promising choices, i.e. choices for which the robot is more likely to be able to push the object closer to the goal. 

For the node $q^{near}$, the weights can be determined by computing the reachability to the goal state $q_u^{goal}$ for each node defined in \ref{sub:distancereachability}. 
For the link number $n_a$, the weights can be calculated based on the distance between the robot and the object centroid. These weights need to be designed for each system depending on the robot and object sizes. 

\subsubsection{Contact Planning}
\label{sub:contactplanning}

The contact planning decides the contact location $p_a$ on the robot and $p_u$ on the object, parameterized respectively by $\phi_a^{(0)}$ and $\phi_u^{(0)}$, as well as the robot configuration $q_a^o$ that realises the desired contact with the object. This is achieved through continuous optimization using our proposed contact location representation explained in \ref{sub:contactlocation}.
In this optimization, described by Eq.\eqref{eq:cp_optimization}
\begin{subequations}
\begin{align}
    \min\limits_{u_u^t, \phi_u^{(0)}, \phi_a^{(0)}, q_a^{(0)}} & 
        \| q_a^{(0)} \|_{W_a} + 
        \sum\limits_{k=1}^{N_{c}}{\| q_u^{(k)} - q_u^f \|_{W_u^k}} \\
    \text{s.t.} \quad
        & q_u^{(k+1)} = q_u^{(k)} + B_u^{(k)} u_u^{(k)} \label{eq:cp_dynamics} \\
        & u^{(k)} \in \mathcal{U}_{B} \cap \mathcal{U}_{FC} \\
        & q^{(0)} \in \mathcal{Q}_{B} \cap \mathcal{Q}_{CF} \cap \mathcal{Q}_{IC} \\
        & \lambda_t = 0, v_u = 0, u_a = 0 \\
        & \| q_u^{(N_c)} - q_u^f \|_{W_u} \leq \| q_u^{(0)} - q_u^f \|_{W_u} \label{eq:cp_improve} \\
        & \Tilde{u}_u (q_u^f, q_u^{(N_c)}) \in \mathcal{U}_{FC} \label{eq:cp_utildeincone},
\end{align}
\label{eq:cp_optimization}%
\end{subequations}
the robot kinematic constraints are enforced only for a first contacting state, while a short-horizon free pushing trajectory from that state is also computed but without robot constraints, and serves to estimate the goodness of the contact location.
Here, $q^{lb}$ and $q^{ub}$ denote the lower and upper boundaries of the state vector $q$ described in \ref{sub:statespacerepr}. The weighted norm $\| \ \|_W$ is defined in Eq.\eqref{eq:weightednorm}. The object dynamics \eqref{eq:cp_dynamics} are detailed in \ref{sub:statespacerepr}. The set $\mathcal{U}_{B}$ corresponds to the control boundaries and $\mathcal{U}_{FC}$ to the contact friction cone (see \ref{sub:contactmodes}). The set $\mathcal{Q}_{CF}$ encodes a collision-free constraint explained in \ref{sub:collisionavoidance} and $\mathcal{Q}_{IC}$ guarantees the robot and the object to be in contact (see \ref{sub:contactlocation}). The final two constraints \eqref{eq:cp_improve} and \eqref{eq:cp_utildeincone} ensure that the final object state $q_u^{(N_c)}$ of the trajectory is closer to the goal state $q_u^{goal}$ than the starting state $q_u^{(0)}$ and that the goal state is still reachable as defined in \ref{sub:distancereachability}.

To keep the optimization time reasonable, the horizon of the trajectory is kept short at this stage. Here we consider less than 10 steps as short-horizon and more than 20 steps as long-horizon. Furthermore, the computation time is decreased by reducing the control trajectory unknown $u_u^t$ to a smaller number of variables and interpolating linearly between them.

\subsubsection{Long-Horizon Guide}
\label{sub:longhorizonguide}

The \textit{Long-Horizon Guide} generates two guiding trajectories in task space: (A) A contact-free trajectory $p_a^t$ that brings the robot target contact location $p_a$ to reach the target contact location on the object $p_u$ while avoiding collision with the object; (B) An in-contact trajectory $q_u^t$ that brings the object closer to the target state $q_u^{goal}$ using TO with a free pusher model. This free pusher abstracts the robot constraints away and aims at capturing the long-horizon behavior of the object dynamics and planning useful sliding on the object surface.

The contact-free trajectory (A) $p_a^t$ can be generated easily by making a trajectory parallel to the object outline function $p_u(\phi_u)$ with some margin distance $D_c(t)$ to minimize the risk of collision.

The in-contact trajectory (B) can be found by solving the TO formulated as
\begin{subequations}
\begin{align}
    \min\limits_{u_u^t} & 
     \sum\limits_{k=1}^{N_{l}}{\| q_u^{(k)} - q_u^f \|_{W_u^k}} \\
    \text{s.t.} \quad
        & q_u^{(k+1)} = q_u^{(k)} + B_u^{(k)} u_u^{(k)} \\
        & u^{(k)} \in \mathcal{U}_{B} \cap \mathcal{U}_{FC} \cap \mathcal{U}_{CC} \\
        & \phi_u^{lb} \leq \phi_u^{(k)} \leq \phi_u^{ub} \label{eq:phiconstraint} \\
        & \| q_u^{(N_{l})} - q_u^f \|_{W_u} \leq \| q_u^{(0)} - q_u^f \|_{W_u},
\end{align}
\label{eq:longhorguide}%
\end{subequations}
where the set $\mathcal{U}_{CC}$ represents the Complementarity Constraint for transition between contact modes defined in \ref{sub:contactmodes}. The constraint \eqref{eq:phiconstraint} can be used to prevent the robot to slide clockwise and counter-clockwise alternatively, which is useful to explore both directions and avoid converging always to the same local solution.

\subsubsection{Guide Tracking}
\label{sub:guidetracking}
The \textit{Guide Tracking} attempts to follow the guide trajectories generated by the \textit{Long-Horizon Guide} while enforcing all the robot kinematic constraints.

For the contact-free trajectory (A), with $\hat{p}_a^{(j)}$ representing the $j^{th}$ point of the guide, the tracking is achieved by solving at every step $k$ the following optimization problem 
\begin{subequations}
\begin{align}
    \min\limits_{u_a^{(k)}} \quad &
       \sum\limits_{j=1}^{N_s} {\| p_a(q_a^{(k+1)}, n_a) - \hat{p}_a^{(j)} \|_{W_a^k}} \\
    \text{s.t.} \quad
        & q^{(k+1)} = q^{(k)} + B^{(k)} u^{(k)} \\
        & u^{(k)} \in \mathcal{U}_B \\
        & q^{(k+1)} \in \mathcal{Q}_B \cap \mathcal{Q}_{CF} \\
        & u_u = 0.
\end{align}
\label{eq:track_contactfree}%
\end{subequations}
Similarly, with $\hat{q}_u^{(j)}$ denoting the $j^{th}$ point of the guide, the tracking of the in-contact trajectory (B) is derived from
\begin{subequations}
\begin{align}
    \min\limits_{u^{(k)}} \quad &
        \sum\limits_{j=1}^{N_s} {\| q_u^{(k+1)} - \hat{q}_u^{(j)} \|_{W^k}} \\
    \text{s.t.} \quad
        & q^{(k+1)} = q^{(k)} + B^{(k)} u^{(k)} \\
        & u^{(k)} \in \mathcal{U}_{B} \cap \mathcal{U}_{FC} \cap \mathcal{U}_{CC} \label{eq:gt_uset} \\
        & q^{(k)} \in \mathcal{Q}_B \cap \mathcal{Q}_{CF} \cap \mathcal{Q}_{IC}.
\end{align}
\label{eq:guidetracking}%
\end{subequations}
At each step of the tracking, the output state $q^{(k+1)}$ state is appended to the tree memory as a new node connected to the node corresponding to state $q^{(k)}$.

The tracking stops under different conditions: reaching the end of the guide trajectory, non-convergence of the optimization problem, obtaining a static solution $u = 0$, or detecting an unexpected collision.

The full planning stops if the last node appended to the tree is close enough to the final goal $q_u^{goal}$ under the distance defined in Eq.\eqref{eq:weightednorm}. Once the final goal has been reached, the shortest path from the initial node to the goal node is extracted with a Dijkstra's algorithm \cite{Dijkstra1959}.

\subsection{Representation}

\subsubsection{Contact Location}
\label{sub:contactlocation}
Our method relies on a novel explicit representation of the contact locations on both the object surface and the robot surface, which in a planar setting reduces to a 2D outline. We represent that location as a parametric function $p(\phi)=(x(\phi), y(\phi))$ with parameter $\phi$.
To build this function, we first process the mesh with an Isotropic Explicit Remeshing using Meshlab \cite{5204086} to improve the uniformity of the vertex density and project all the vertices of the mesh to the plane $(e_x, e_y)$. We then extract and discretize the 2D outline of the shape (for e.g., using the technique described in \cite{6710113}). With $N_p$ the number of vertices on the outline, $p_n$ the $n^{th}$ vertex and $\bar{p}$ the centroid of all vertices, the fitting function is defined as 
\begin{equation}
    p(\phi) = \bar{p} + \sum_{n=0}^{N_p} b_n(\phi) (p_n - \bar{p}),
    \label{eq:gaussianfit}%
\end{equation}
where $b_n(\phi)$ are coefficient functions that we chose to be Gaussians centered on the value of $\phi_n=n/N_p$ with variance $\sigma^2={(1 / N_p)}^2$, as 
\begin{equation}
    b_n(\phi) = \frac{1}{N_p \sigma \sqrt{\pi}} \exp{(-(\frac{\phi-\phi_n}{\sigma})^2)}.
    \label{eq:gaussiancoef}%
\end{equation}
Consequently, noticing that the position of the vertices depend on the robot state $q_a$ and the object state $q_u$, the in-contact condition of the robot and the object becomes
\begin{equation}
    \mathcal{Q}_{IC} = \{ q \in {\rm I\!R}^8 | \ p_u(q_u) = p_a(q_a) \}.
    \label{eq:q_ic}%
\end{equation}

\subsubsection{State-Space Representation}
\label{sub:statespacerepr}
Similar to \cite{moura2022nonprehensileplanarmanipulationtrajectory}, we adopt the linear state-space representation
\begin{equation}
    q^{(k+1)} = q^{(k)} + B^{(k)} u^{(k)}. 
    \label{eq:statespace}%
\end{equation}
Here, $k$ denotes the timestep number, the input matrix $B$ is described in Eq.\eqref{eq:inputmatrix} and the state vector $q$ is defined as $q=(q_u^\top, q_a^\top)^\top$ with $q_u=(x_u, y_u, \alpha_u, \phi_u)^\top$ and $q_a=(\theta_1, \theta_2, \theta_3, \phi_a)^\top$, 
where $x_u, y_u$ are the position coordinates of the object, $\alpha_u$ is its orientation, $\theta_1, \theta_2, \theta_3$ are the joint angles of the robot and $\phi_u$ and $\phi_a$ are the parameters for the contact location on the object and the robot respectively (see \ref{sub:contactlocation}).
Also, $u$ is the control vector defined as $u=(u_u^\top, u_a^\top)^\top$ with $u_u=(\lambda_n, \lambda_t, v_u)^\top$ and $u_a=(\tau_1, \tau_2, \tau_3, v_a)^\top$, 
$\lambda_n$ and $\lambda_t$ being the normal and tangential components of the contact impulse $\lambda_u$, $\tau_1, \tau_2, \tau_3$ the torque for each robot joints and $v_u$ and $v_a$ the sliding velocity on the object and robot outlines, respectively.
Boundaries are set on the state and control vectors as
\begin{subequations}
\begin{align}
    &\mathcal{Q}_{B} = \{ q \in {\rm I\!R}^8 | \ q^{lb} \leq q \leq q^{ub} \} \\
    &\mathcal{U}_{B} = \{ u \in {\rm I\!R}^7 | \ u^{lb} \leq u \leq u^{ub} \}.
\end{align}
\label{eq:boundaries}%
\end{subequations}

For the object dynamics, we use the same ellipsoidal approximation of the limit surface model as in \cite{hogan2020reactive} which is formulated as follows.
\begin{subequations}
\begin{align}
    & q_u^{(k+1)} = q_u^{(k)} + B_u^{(k)} u_u^{(k)} \\
    \text{with} \quad
        & B_u = \begin{pmatrix}
            - E_p^\top L J_c^\top & 0_{3\times1} \\
            0_{1\times2} & 1
            \end{pmatrix} \\
        & E_p = \begin{pmatrix}
            e_x          & e_y          & 0_{3\times1} \\
            0_{3\times1} & 0_{3\times1} & e_z          \\
            \end{pmatrix}\\
        & L = \begin{pmatrix}
            1_{3\times3} & 0_{3\times3} \\
            0_{3\times3} & k_L 1_{3\times3} \\
            \end{pmatrix},
\end{align}
\label{eq:qudynamics}%
\end{subequations}
where $k_L$ is the limit surface coefficient, $J_c \in {\rm I\!R}^{2\times6}$ the Jacobian matrix of the contact normal $e_n$ and tangent $e_t$ at the object contact location. This model allows to add more contact points by simply replacing $J_c^{\top}$ with a sum $\sum\limits_{i=1}^{N}{J_{ci}^{\top}}$ on $N$ contact impulses and extending the variables $\lambda_n$, $\lambda_t$, $\phi_u$, $\phi_a$ and $n_a$ accordingly.

For the robot dynamics, a multi-body system dynamics has the following general form \cite{Liu2012AQT} 
\begin{equation}
    M(\theta) \ddot{\theta} + C(\theta, \dot{\theta}) = J_v^\top(\theta) f + J_\omega^\top(\theta) \tau,
    \label{eq:fulldynamics}%
\end{equation}
where $\theta$ is the joint angles vector, $M$ stands for the mass matrix, $C$ for the Coriolis and centrifugal term, $f$ and $\tau$ respectively the forces and torques applying on the robot and $J_v$ and $J_\omega$ the corresponding Jacobian matrices of the location where they apply. Under quasi-dynamic assumption, we approximate the left-hand side of Eq.\eqref{eq:fulldynamics} with
\begin{equation}
    M(\theta) \ddot{\theta} + C(\theta, \dot{\theta}) \approx \epsilon \frac{\Delta \theta}{\Delta t},
    \label{eq:quasidynamics}%
\end{equation}
$\Delta t$ being the time step of the planning and $\epsilon$ a regularization term that we choose to be $\epsilon = \Delta t$. 

Replacing \eqref{eq:quasidynamics} in Eq.\eqref{eq:fulldynamics}, we get the state-space update equation for the robot
\begin{subequations}
\begin{align}
    & q_a^{(k+1)} = q_a^{(k)} + B_a u_a^{(k)} + H_a^{(k)} u_u^{(k)} \\
    \text{with} \quad
        & B_a = \begin{pmatrix}
            \frac{\Delta t}{\epsilon} 1_{3\times3} & 0_{3\times1} \\
            0_{1\times3} & 1 \\
            \end{pmatrix}\\
        & H_a = -\frac{1}{\epsilon} \begin{pmatrix}
            X_1 & 0 \\  
            X_2 & 0 \\ 
            X_3 & 0 \\
            0_{1\times2} & 0 \\
            \end{pmatrix} \\
        & X_i = \left \{ \begin{array}{l}
            e_z^\top [p_c - p_i] \begin{pmatrix} e_n & e_t \end{pmatrix} \quad \text{if} \ i \leq {n_a} \\
            0_{1\times2} \qquad \qquad \qquad \quad \; \; \; \text{otherwise}
        \end{array} \right. ,
\end{align}
\label{eq:qadynamics}%
\end{subequations}
where $p_c$ is the position where the contact force applies, $p_i$ is the position of the $i^{th}$ joint origin and $[ p ]$ defines the skew-symmetric matrix of the cross product $p \cdot$.

Putting together equations \eqref{eq:statespace}, \eqref{eq:qudynamics} and \eqref{eq:qadynamics} yields
\begin{equation}
    B = \begin{pmatrix} 
        B_u & 0_{4\times4} \\
        H_a & B_a  \\
    \end{pmatrix}.
    \label{eq:inputmatrix}%
\end{equation}

\subsubsection{Collision avoidance}
\label{sub:collisionavoidance}
\begin{figure}[t]
    \centering
    \includegraphics[width=0.70\linewidth, trim=0 5 2 2, clip]{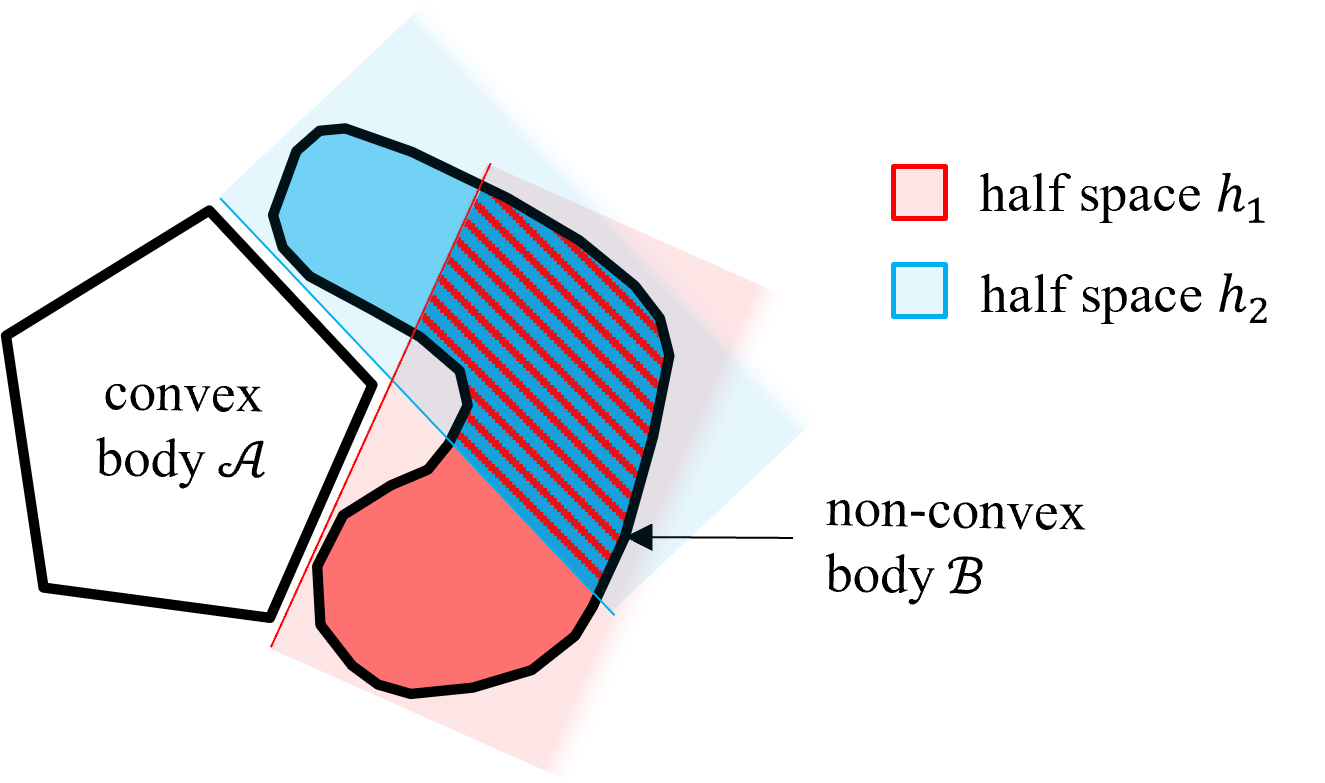}
    \captionsetup{justification=centering,margin=0.2cm}
    \caption{Half-space separation between one convex body and one non-convex body to determine collision}
    \label{fig:collisionavoidance}
\end{figure}

To avoid undesired collision between the robot and the object, we express the collision avoidance as a constraint in our optimization programs.
Since the shape of the robot links is, in general, non-convex, we extend the formulation for collision avoidance between two convex bodies using half-space separation plane \cite{8246920} \cite{amice2024certifyingbimanualrrtmotion} to the special case of one convex body $\mathcal{A}$ and one non-convex body $\mathcal{B}$. As shown in Fig.\ref{fig:collisionavoidance}, for each vertex $p$ on the outline of the non-convex body $\mathcal{B}$, if a half-space that includes $p$ and does not intersect with the convex body $\mathcal{A}$ can be found, then there is no collision. Considering that the position of the non-convex links depends on the robot configuration $q_a$ and the half-spaces depend on the convex object state $q_u$, the collision-free condition can then be formulated as
\begin{equation}
    \begin{split}
    \mathcal{Q}_{CF} = &\{ q \in {\rm I\!R}^8 | \\
        &\min\limits_{p \in \mathcal{B}} {(\max\limits_{h \in \mathcal{H}}{(\ n_h^\top(q_u) (p(q_a) - p_h(q_u)))} )} > 0 \},
    \end{split}
    \label{eq:q_cf}%
\end{equation}
where $\mathcal{H}$ is a set of half-space separation planes $h$ defined by an origin $p_h$ and a normal $n_h$. We approximate the object and robot shapes as polygons to limit the problem to a finite number of half-spaces.

Collision between the different links of the robot that cannot be enforced by a constraint on the joint angles can be checked after the optimization using any off-the-shelf algorithm before adding a new node to the tree.

\definecolor{darkgreen}{RGB}{25,200,25}

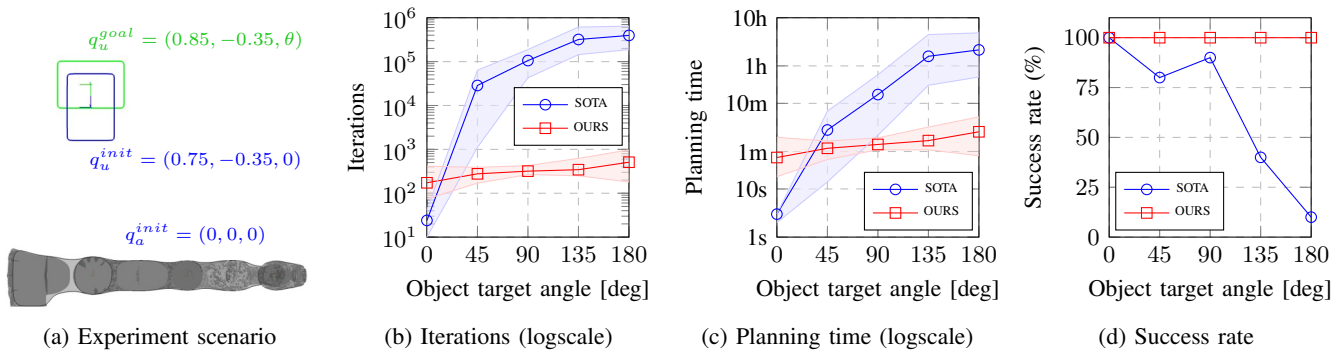
\begin{figure*}[t]
    \centering
    \begin{subfigure}[b]{0.24\textwidth}
        \scriptsize
        \begin{tikzpicture} 
            \node {\includegraphics[width=0.95\linewidth, trim={0 1.5cm 0 0}]{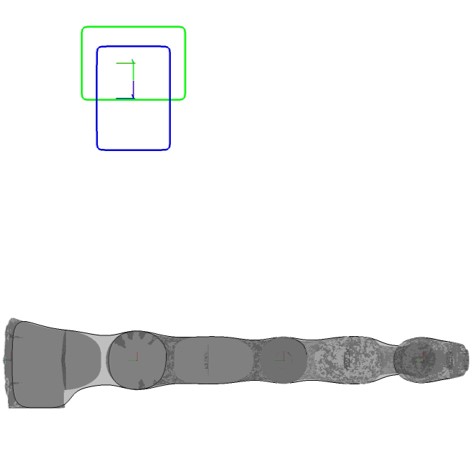}};

            \node at (0.5, -0.8) [color=blue] {$q_a^{init}=(0, 0, 0)$};
            \node at (0.5, 0.2) [color=blue] {$q_u^{init}=(0.75, -0.35, 0)$};
            \node at (0.5, 1.8) [color=darkgreen] {$q_u^{goal}=(0.85, -0.35, \theta)$};
        
        \end{tikzpicture}
        
        \centering
        \captionsetup{justification=centering,margin=0.0cm}
        \caption{Experiment scenario}
        \label{fig:scenario1}
    \end{subfigure}
    \hfill
    \centering
    \begin{subfigure}[b]{0.24\textwidth}
        \begin{tikzpicture} 
        \centering
        \begin{semilogyaxis}[
            xlabel={\small{Object target angle [deg]}},
            ylabel={\small{Iterations}},
            xmin=0, xmax=180, ymin=10, ymax=1000000,
            xtick={0,45,90,135,180},
            ytick={10,100,1000,10000,100000, 1000000},
            legend style={at={(0.43,0.55)},anchor=west},
            ymajorgrids=true, yminorgrids=false, xmajorgrids=true,
            height=4.5cm,
            grid style=dashed, 
            ticklabel style = {font=\small},
            width=1\textwidth]
    
        \addplot[color=blue, mark=o]
            coordinates { (0,24)(45,28599)(90,105701)(135,319778)(180,395395) };
        \addplot[color=red, mark=square]
            coordinates { (0,173)(45,277)(90, 319)(135,344)(180,509) };
            
        \addplot[color=blue!20, name path=min1]
            coordinates { (0,10)(45,1111)(90,42759)(135,144168)(180,188196) };
        \addplot[color=blue!20, name path=max1]
            coordinates { (0,41)(45,63642)(90,188458)(135,613157)(180,644639) };
        \addplot[fill=blue!10, fill opacity=0.6] fill between[of=min1 and max1];
        
        \addplot[color=red!20, name path=min2]
            coordinates { (0,74)(45,170)(90,266)(135,250)(180,181) };
        \addplot[color=red!20, name path=max2]
            coordinates { (0,414)(45,389)(90,426)(135,623)(180,972) };
        \addplot[fill=red!10, fill opacity=0.6] fill between[of=min2 and max2];
        
        \legend{\tiny{SOTA}, \tiny{OURS}}
        \end{semilogyaxis} \end{tikzpicture}
        \centering
        \captionsetup{justification=centering,margin=0.0cm}
        \caption{Iterations (logscale)} 
        \label{fig:resultnodes}
    \end{subfigure}
    \hfill
    \centering
    \begin{subfigure}[b]{0.24\textwidth}
        \begin{tikzpicture} 
        \centering
        \begin{semilogyaxis}[
            xlabel={\small{Object target angle [deg]}},
            ylabel={\small{Planning time}},
            xmin=0, xmax=180, ymin=1, ymax=3600*10,
            xtick={0,45,90,135,180},
            ytick={1, 10, 60, 600, 3600, 3600*10},
            yticklabels={1s, 10s, 1m, 10m, 1h, 10h},
            legend pos=south east,
            ymajorgrids=true, yminorgrids=true, xmajorgrids=true,
            height=4.5cm,
            grid style=dashed, 
            ticklabel style = {font=\small},
            width=1\textwidth,
            ]
        \addplot[color=blue, mark=o]
            coordinates { (0,3)(45,168)(90,914)(135,5700)(180,7742) };
        \addplot[color=red, mark=square]
            coordinates { (0,45)(45,70)(90, 84)(135,101)(180,154) };
            
        \addplot[color=blue!20, name path=min1]
            coordinates { (0,2)(45,14)(90,130)(135,1427)(180,2116) };
        \addplot[color=blue!20, name path=max1]
            coordinates { (0,3)(45,413)(90,2364)(135,16078)(180,17467) };
        \addplot[fill=blue!10, fill opacity=0.6] fill between[of=min1 and max1];
        
        \addplot[color=red!20, name path=min2]
            coordinates { (0,18)(45,41)(90,65)(135,65)(180,49) };
        \addplot[color=red!20, name path=max2]
            coordinates { (0,118)(45,99)(90,115)(135,193)(180,314) };
        \addplot[fill=red!10, fill opacity=0.6] fill between[of=min2 and max2];
        
        \legend{\tiny{SOTA}, \tiny{OURS}}
        \end{semilogyaxis} \end{tikzpicture}
        \centering
        \captionsetup{justification=centering,margin=0.0cm}
        \caption{Planning time (logscale)} 
        \label{fig:resulttime}
    \end{subfigure}
    \hfill
    \centering
    \begin{subfigure}[b]{0.24\textwidth}
        \begin{tikzpicture} 
        \centering
        \begin{axis}[
            xtick={0, 45, 90, 135, 180},
            ytick={0, 25, 50, 75, 100},
            xmin=0, xmax=180, ymin=0, ymax=110,
            xlabel={\small{Object target angle [deg]}},
        	ylabel={\small{Success rate (\%)}},
        	legend pos=south west,
            ymajorgrids=true, yminorgrids=true, xmajorgrids=true,
            height=4.5cm,
            grid style=dashed, 
            ticklabel style = {font=\small},
            width=1\textwidth]
        \addplot[color=blue, mark=o]
        	coordinates {(0,100) (45,80) (90,90) (135,40) (180, 10)};
        \addplot[color=red, mark=square]
        	coordinates {(0,100) (45,100) (90,100) (135,100) (180, 100)};
        \legend{\tiny{SOTA}, \tiny{OURS}}
        \end{axis} 
        \end{tikzpicture}
        \centering
        \captionsetup{justification=centering,margin=0.0cm}
        \caption{Success rate} 
        \label{fig:resultsuccess}
    \end{subfigure}
    \centering
    \captionsetup{justification=centering,margin=1.0cm}
    \caption{Planning performance comparison between the state-of-the-art (SOTA) and our method (OURS).}
    \label{fig:resconvergence}
\end{figure*}

\subsubsection{Contact modes as Complementarity Constraints}
\label{sub:contactmodes}

We define the set $\mathcal{U}_{FC}$ of contact impulses that lie in the friction cone derived from Coulomb friction \cite{174511} at the contact point and the set $\mathcal{U}_{CC}$ that encodes the transition between sticking and sliding contact modes as a Complementarity Constraint similar to \cite{posa2014directmethod}. In our definition of $\mathcal{U}_{CC}$, we account for the sliding that may occur on both the object surface and the robot surface, formulated as
\begin{subequations}
\begin{align}
    &\mathcal{U}_{FC} = \{ u \in {\rm I\!R}^7 | 
        -\mu \lambda_n \leq \lambda_t \leq \mu \lambda_n, 
        \lambda_n \geq 0 \} \\
    \begin{split}
    \mathcal{U}_{CC} = \{ u \in {\rm I\!R}^7 |
        (v_u - v_a) (\lambda_t^2 - \mu^2 \lambda_n^2) = 0, \\
        v_u \lambda_t \geq 0, 
        v_u v_a \leq 0 \},
    \end{split}
\end{align}
\label{eq:constraintsets}%
\end{subequations}
where $\mu$ is the friction coefficient of the contact between the robot and the object.

\subsubsection{State Distance and Reachability}
\label{sub:distancereachability}
We define the distance between two states as a weighted norm of the difference $\delta q$ between two state vectors, with $W$ the diagonal matrix of the weights for each component of the state
\begin{equation}
    \| \delta q \|_W = {(W {\delta q})}^\top (W \delta q).
    \label{eq:weightednorm}%
\end{equation}
We also define a reachability metric which measures how easy it is to reach a target object state $q_u^f$ from another state $q_u$, based on the input matrix $B_u$ defined in Eq.\eqref{eq:statespace}.
\begin{subequations}
\begin{align}
    & \mathcal{R}(q_u^f, q_u) = \exp{(-\Tilde{u}_u - \gamma(\Tilde{u}_u))} \\
    \text{with} \quad
        & \Tilde{u}_u (q_u^f, q_u) = {B_u(q_u)}^+ (q_u^f - q_u) \\
        & \gamma (\Tilde{u}_u) = 0 \ \text{if} \ \Tilde{u}_u \in \mathcal{U}_{FC}, 
        \gamma_0 > 0 \ \text{otherwise}         
\end{align}
\label{eq:reachability}%
\end{subequations}
where $B_u^+$ denotes the pseudoinverse $B_u^+ = {(B_u^\top B_u)}^{-1} B_u^\top$.
A penalty term $\gamma_0$ is used to penalize states that require an impulse that is outside of the friction cone.

\section{RESULTS}

We evaluate our method through four different experiments.
Experiment 1 compares the performance of our planning with the state-of-the-art sampling method. 
Experiment 2 evaluates the planning performance with and without some of the main components of our pipeline to verify their benefits. 
Experiment 3 assesses the versatility of the proposed method with contact and motion planning using different links of the robot and different object shapes. 
Experiment 4 aims at checking the precision of the contact location planned with our method by demonstrating it on a real hardware.

The simulation experiments 1-3 have been conducted on a desktop computer with the following features: Intel\textregistered \ Core\texttrademark \ i9-9900K CPU @ 3.60GHz x 16, Nvidia Geforce GTX 900M, 64Gb of RAM. The robot considered for all experiments is the KUKA IIWA 14.
Optimization programs in our method are solved using CasADI \cite{Andersson2018CasADiAS} with the solver IPOPT \cite{Wchter2006OnTI} wrapped in the package OpTaS \cite{Mower_2023}.

\subsection{Experiment 1: Comparison with the state-of-the-art}

\subsubsection{Protocol} 
We compare the performance of WBCRM planning of our method with the state-of-the-art (SOTA) sampling method {\small\texttt{global\_planning\_contact}} provided by T. Pang et al. in \cite{pang2023globalplanningcontactrichmanipulation}. We modified the available example iiwa\_box\_push by rotating the robot $90^{\circ}$ to be planar and keeping only the joints whose rotation axis is along $e_z$. The SOTA planning uses NLOPT solver \cite{NLopt} and Drake simulator \cite{drake}.
We tailored a scenario, shown in Fig.\ref{fig:scenario1}, of re-orienting a box (276x198mm size) to different angles, from the start pose $q_u^{init}=(0.75, -0.35, 0)$ to the goal pose $q_u^{goal}=(0.85, -0.35, \theta)$ with $\theta$ taking the values $0^{\circ}, 45^{\circ}, 90^{\circ}, 135^{\circ} \ \text{and} \ 180^{\circ}$ (position in [m]). The goal position is set near the edge of the robot range of motion purposely as it corresponds to a common scenario in which contact-rich manipulation extends the robot reachable space. The difficulty of the planning is expected to increase with the target angle due to the necessity of changing more frequently the pushed face of the box without pushing it out of the workspace of the robot.
For each planning, we measured the number of iterations required to reach the goal pose, the planning time and the success rate over 10 attempts.

\subsubsection{Results} 
As shown in the chart Fig.\ref{fig:resultnodes}, our method greatly improved convergence for pushing that involves re-orientation of the object, reducing the number of iterations needed to reach the goal state by 99\% in average. Similarly, the planning time decreases by 96\% on average: from about 3min to 1min for $\theta=45^{\circ}$, and up to 2h to 3min reduction for $\theta=180^{\circ}$ (Fig.\ref{fig:resulttime}). The increase of time for motion without re-orientation is due to the overhead time induced by solving the different optimizations in our pipeline. 

Moreover, the holistic approach of our planning which implies that all constraints are enforced at each iteration minimizes the risk of infeasibility later on, whereas the state-of-the-art method sometimes failed to plan collision-free trajectories during the trajectory refinement that takes place after the in-contact motion planning, as Fig.\ref{fig:resultsuccess} reveals.

\begin{figure}[t]
    \centering
    \begin{subfigure}[t]{0.23\textwidth}
        \scriptsize
        \begin{tikzpicture} 
            \node {\includegraphics[width=1\linewidth, trim={3cm 2cm 5cm 20cm}, clip]{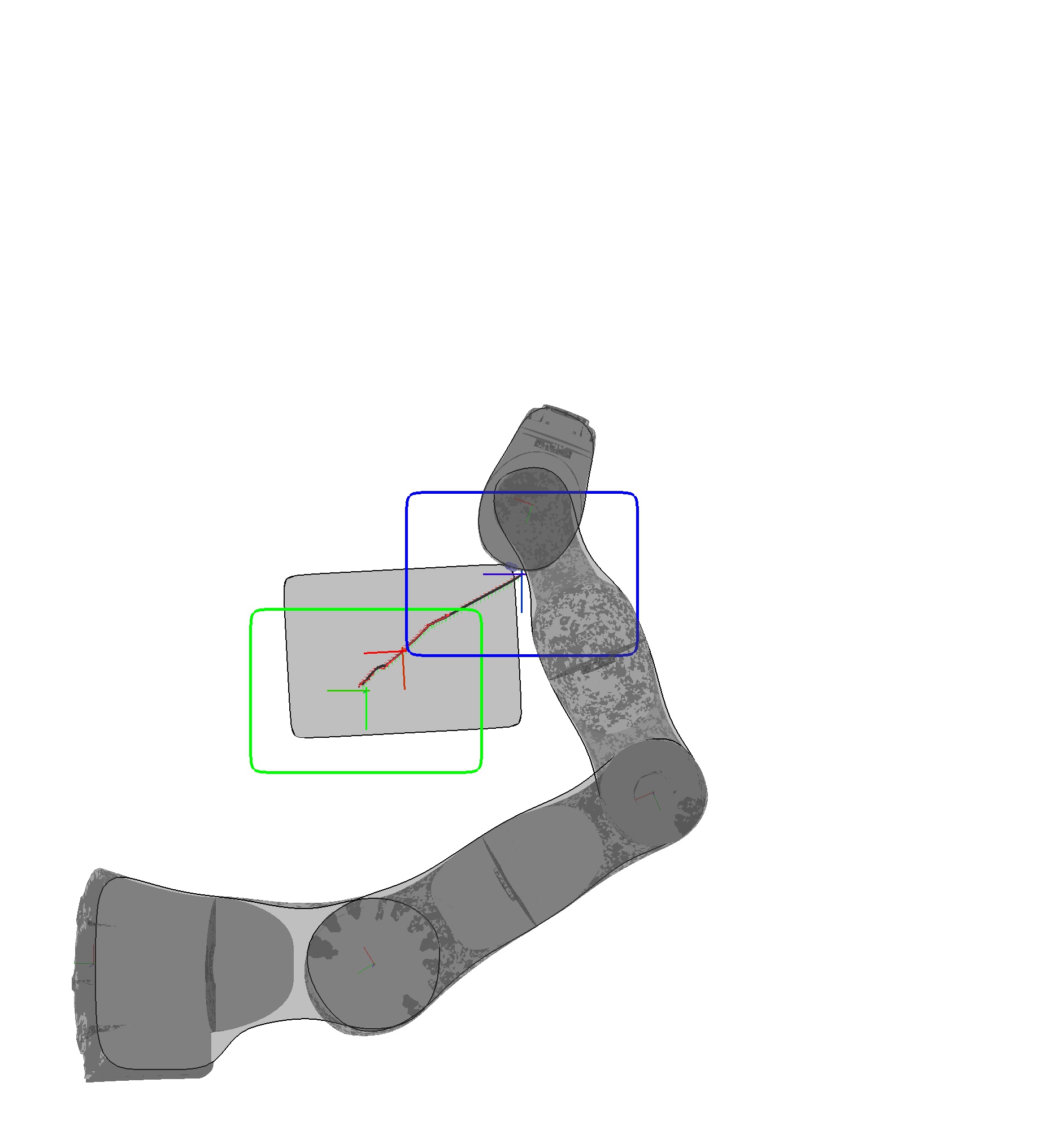}};

            \node at (1.0, 1.0) [color=blue] {$q_u^{init}$};
            \node at (-0.8, -0.5) [color=darkgreen] {$q_u^{goal}$};
            \draw[->] (-0.7, 1.3) node [anchor=south, text centered, align=center] {expected behavior: \\ slide over the corner to \\ efficiently switch face} -- (-0.2, 0.8);
        
        \end{tikzpicture}
        \centering
        \captionsetup{justification=centering,margin=0.0cm}
        \caption{Scenario 1}
        \label{fig:exp2_scenario1}
    \end{subfigure}
    \hfill
    \centering
    \begin{subfigure}[t]{0.23\textwidth} 
        \scriptsize
        \begin{tikzpicture}
            \node {\includegraphics[width=1\linewidth, trim={3cm 2cm 5cm 20cm}, clip]{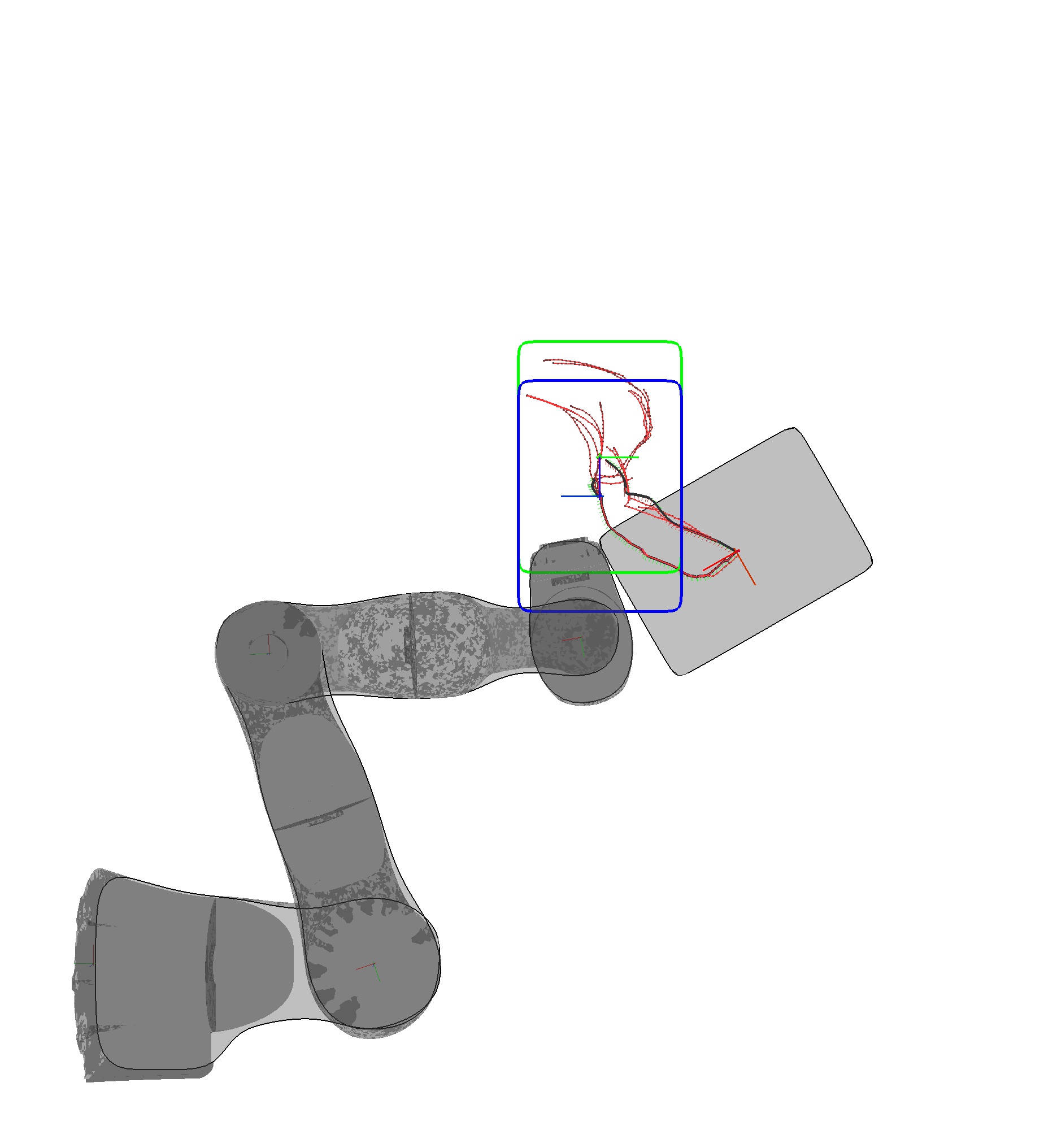}};
    
            \node at (-0.3, 1.0) [color=blue] {$q_u^{init}$};
            \node at (-0.3, 1.8) [color=darkgreen] {$q_u^{goal}$};
            \draw[->] (1.0, -0.5) node [anchor=north, text centered, align=center] {expected behavior: \\ move away to \\ rotate the object} -- (1.0, 0.2);
        \end{tikzpicture} 
        \centering
        \captionsetup{justification=centering,margin=0.0cm}
        \caption{Scenario 2}
        \label{fig:exp2_scenario2}
    \end{subfigure}

    \centering
    \begin{subfigure}[t]{0.45\textwidth} 
    \vspace{0.2cm}
        \centering
        \small
        \begin{tabular}{ |l||c||r||r| }
             \hline
             Planning result& Sce.& Success& Mean time\\
             \hline
             \multirow{2}{*}{Full pipeline}
             & 1& \textbf{100\%}& \textbf{00:02:44}\\
             & 2& \textbf{40\%}& \textbf{00:07:27}\\
             \hline
             \multirow{2}{*}{No \textit{Long-Horizon Guide}}
             & 1& 20\%& 00:24:15\\
             & 2& 0\%& -\\
             \hline
             \multirow{2}{*}{No contact optimization}
             & 1& 20\%& 00:15:43\\
             & 2& 0\%& -\\
             \hline
        \end{tabular}
    \end{subfigure}
    \centering
    \captionsetup{justification=centering,margin=0.1cm}
    \caption{Experiment 2 scenarios overview with planned trajectory with the full pipeline and comparative results}
\end{figure}

\subsection{Experiment 2: Benefit of the Proposed Method}
\subsubsection{Protocol}
To confirm the benefits of the main components of our pipeline, we investigate the impact on the planning performance of the following pipeline variations: a) a pipeline without \textit{Long-Horizon Guide}; b) a pipeline with random contact planning and no sliding on the robot surface. The variation a) is realized by replacing the guiding trajectory $\hat{q}_u^{(j)}$ with the constant goal state $q_u^{goal}$. The variation b) is realized by randomly sampling contact locations and solving the inverse kinematics for that sample, as well as constraining the sliding on the robot to be $v_a=0$. 
We consider the planning of two motion scenarios: a translation scenario 1 with $q_u^{init}=(0.50, -0.55, 90)$ and $q_u^{goal}=(0.35, -0.35, 90)$ pictured in Fig.\ref{fig:exp2_scenario1} and a rotation scenario 2 with $q_u^{init}=(0.65, -0.65, 0)$ and $q_u^{goal}=(0.70, -0.65, 180)$ pictured in Fig.\ref{fig:exp2_scenario2}. For each setting, we collected the planning time and the success rate over 5 attempts, with any planning exceeding a timeout of 1h considered as failure.

\subsubsection{Results}
a) When the \textit{Long-Horizon Guide} is removed, in scenario 1 the robot cannot find within a reasonable time the right spot where to switch pushing face, shown with an arrow on Fig.\ref{fig:exp2_scenario1}, to be able to reach the goal before the object overshoots to the left. In scenario 2, the robot needs first to move the object away from the goal position to be able to rotate it before translating it towards the goal, as the arrow on Fig.\ref{fig:exp2_scenario2} shows. This trajectory could not be found within a reasonable time without the \textit{Long-Horizon Guide}.

b) Without the contact optimization, the robot often contacts on inefficient locations of the object, resulting in losing time trying to push the object from unfavorable positions.

These results prove the crucial role of the components of our pipeline in improving the planning convergence, while the low success rate on scenario 2 reminds that fast planning of long-horizon WBCRM remains a challenging task.

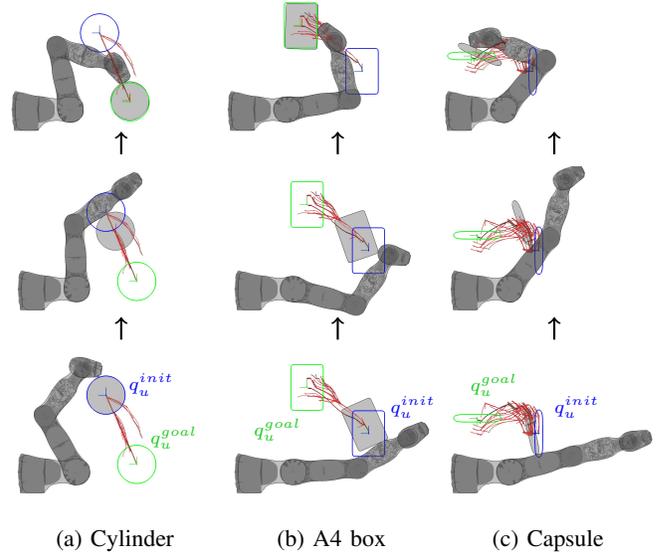
\begin{figure}[t]
    \centering
    \begin{subfigure}[t]{0.32\linewidth} 
        \includegraphics[width=1\linewidth, trim=255 800 255 300, clip]{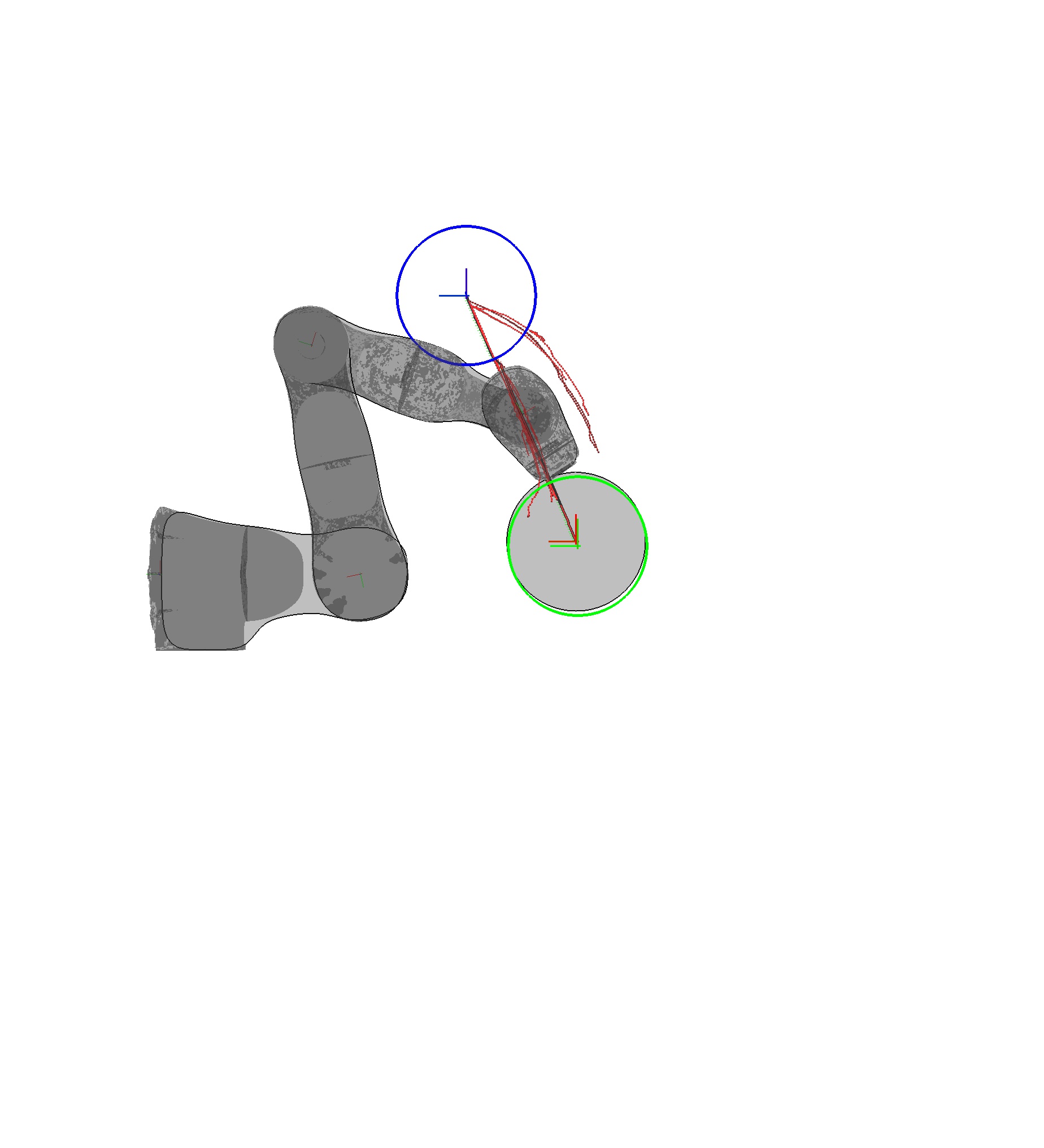} 
    \end{subfigure} 
    \hfill
    \begin{subfigure}[t]{0.32\linewidth} 
        \includegraphics[width=1\linewidth, trim=255 800 255 300, clip]{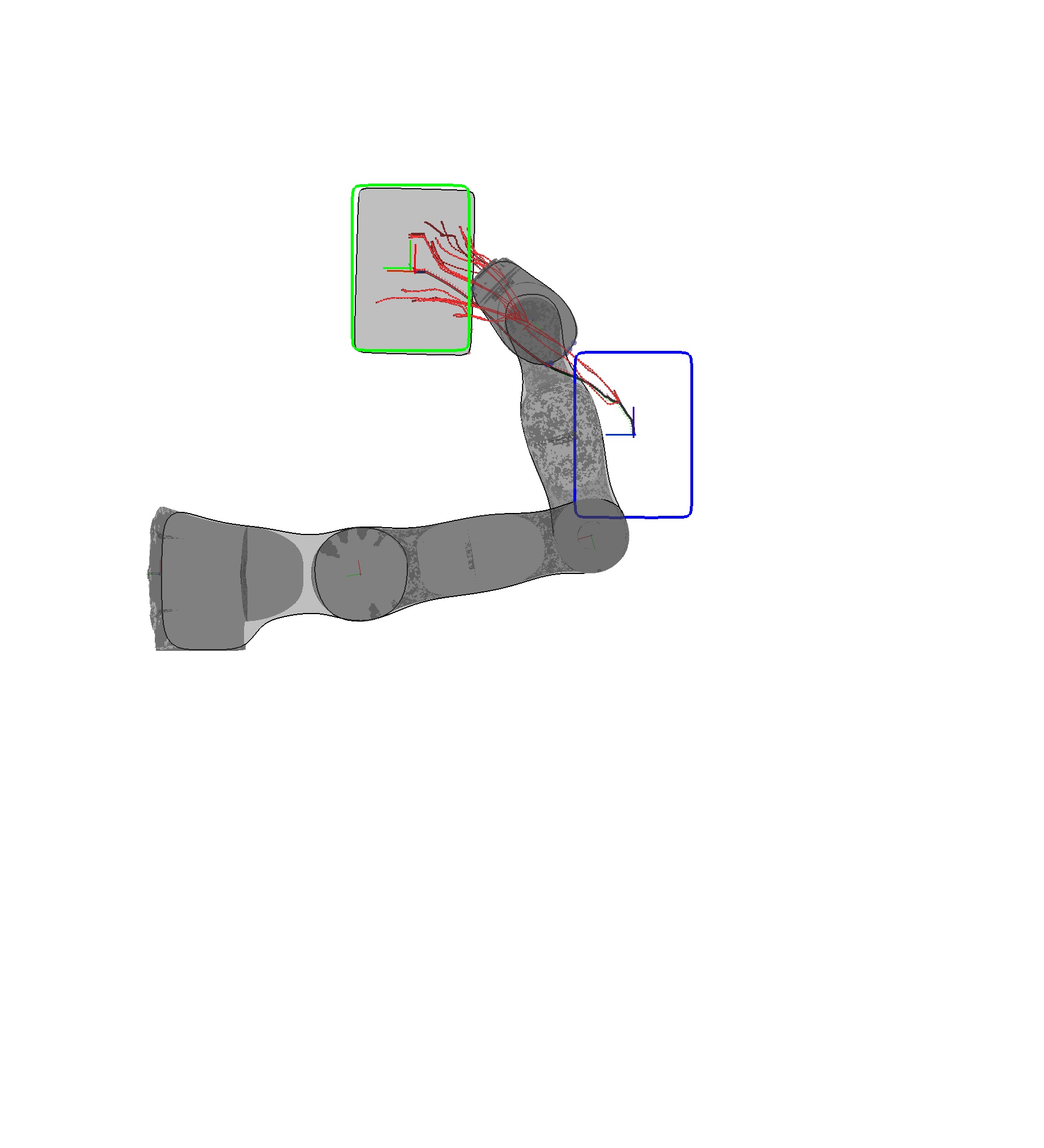} 
    \end{subfigure} 
    \hfill
    \begin{subfigure}[t]{0.32\linewidth} 
        \includegraphics[width=1\linewidth, trim=255 800 255 300, clip]{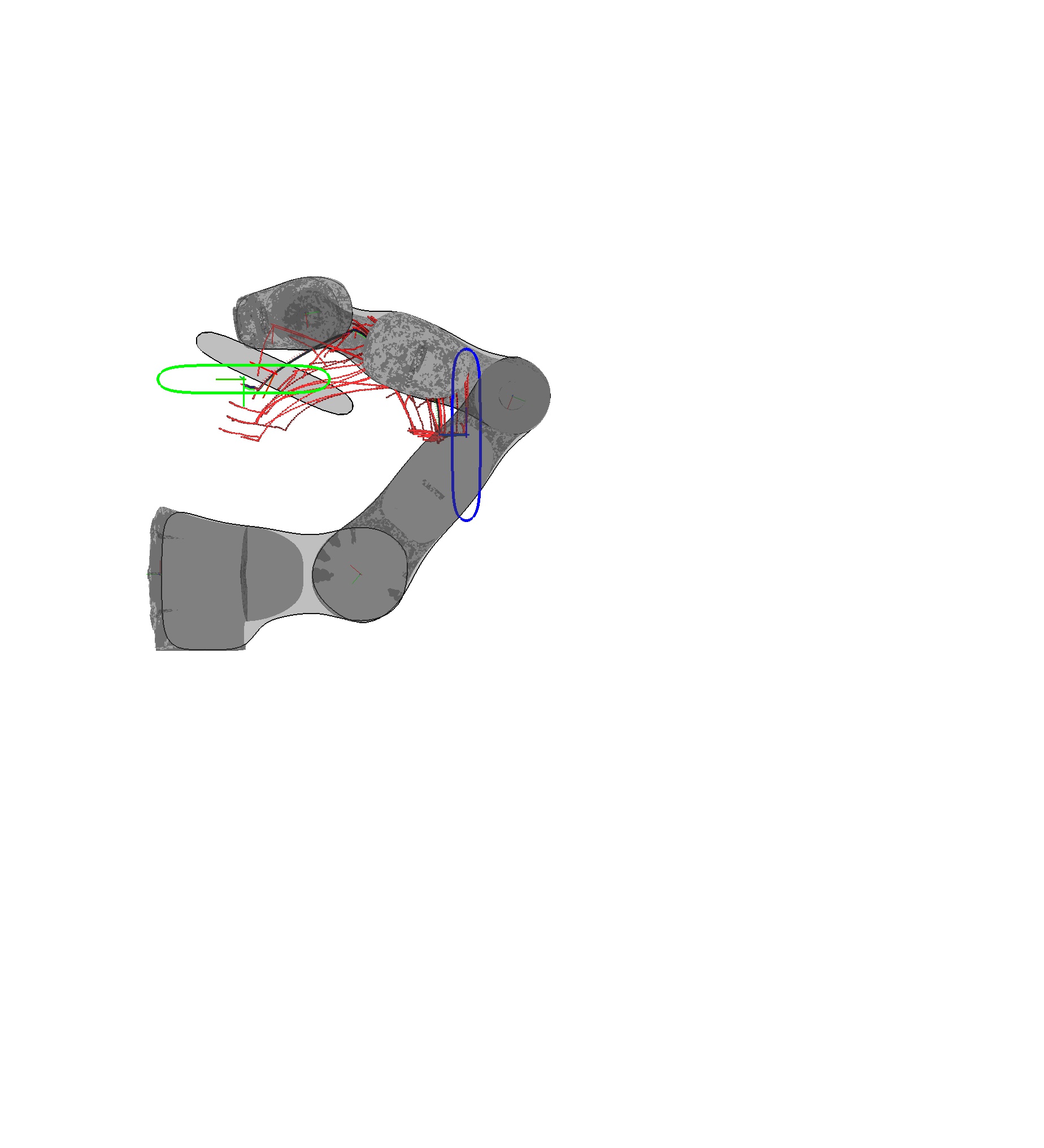}
    \end{subfigure} 
    \hfill \vspace{-0.17cm}

    \hfill
    \begin{subfigure}[t]{0.32\linewidth} 
        \scriptsize
        \begin{tikzpicture} 
            \node {\includegraphics[width=1\linewidth, trim=255 800 255 225, clip] {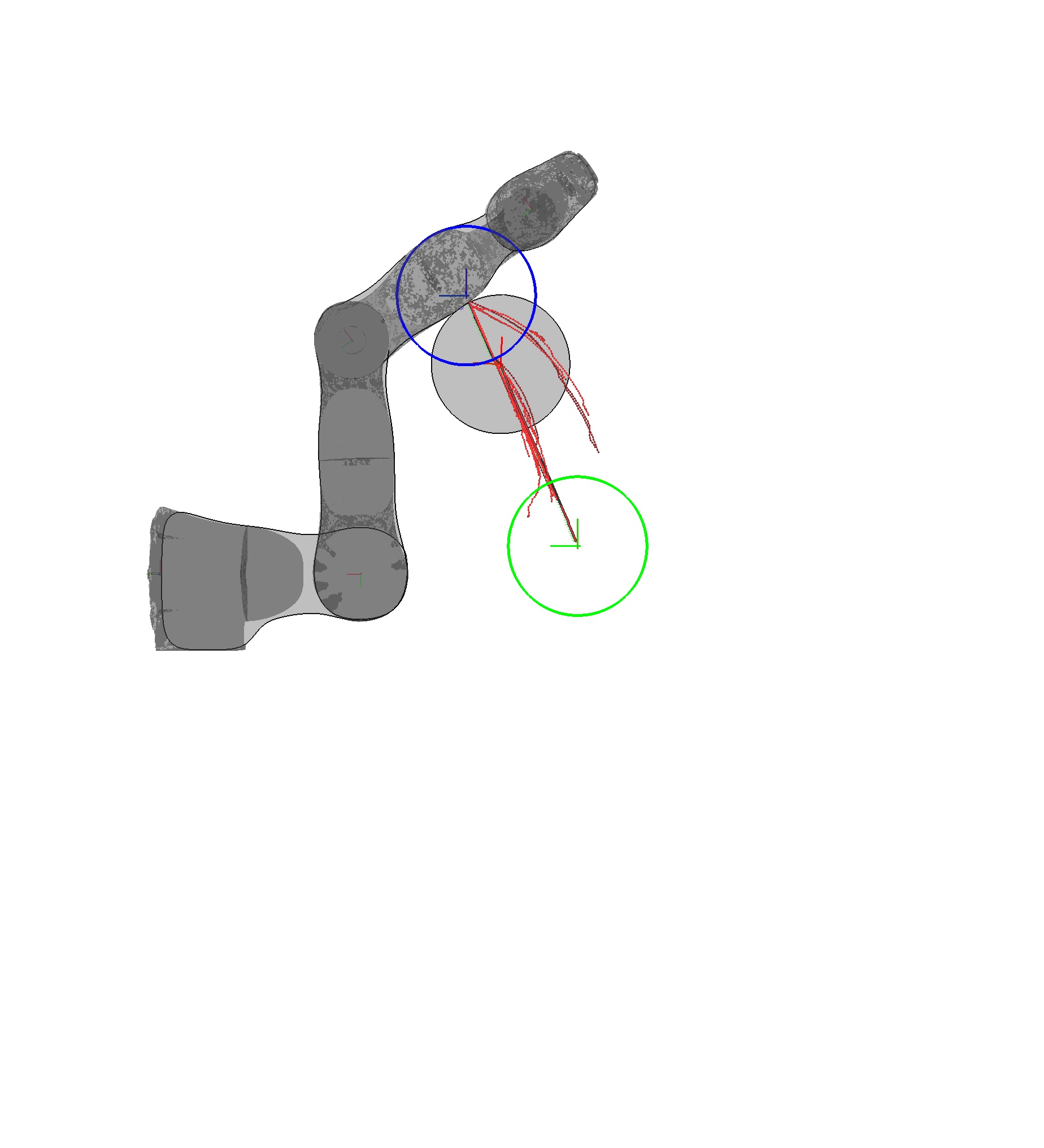}};
            \draw[->, thick] (0.0, 1.2) -- (0.0, 1.5);
        \end{tikzpicture}
    \end{subfigure} 
    \hfill
    \begin{subfigure}[t]{0.32\linewidth} 
        \scriptsize
        \begin{tikzpicture} 
            \node {\includegraphics[width=1\linewidth, trim=255 800 255 225, clip] {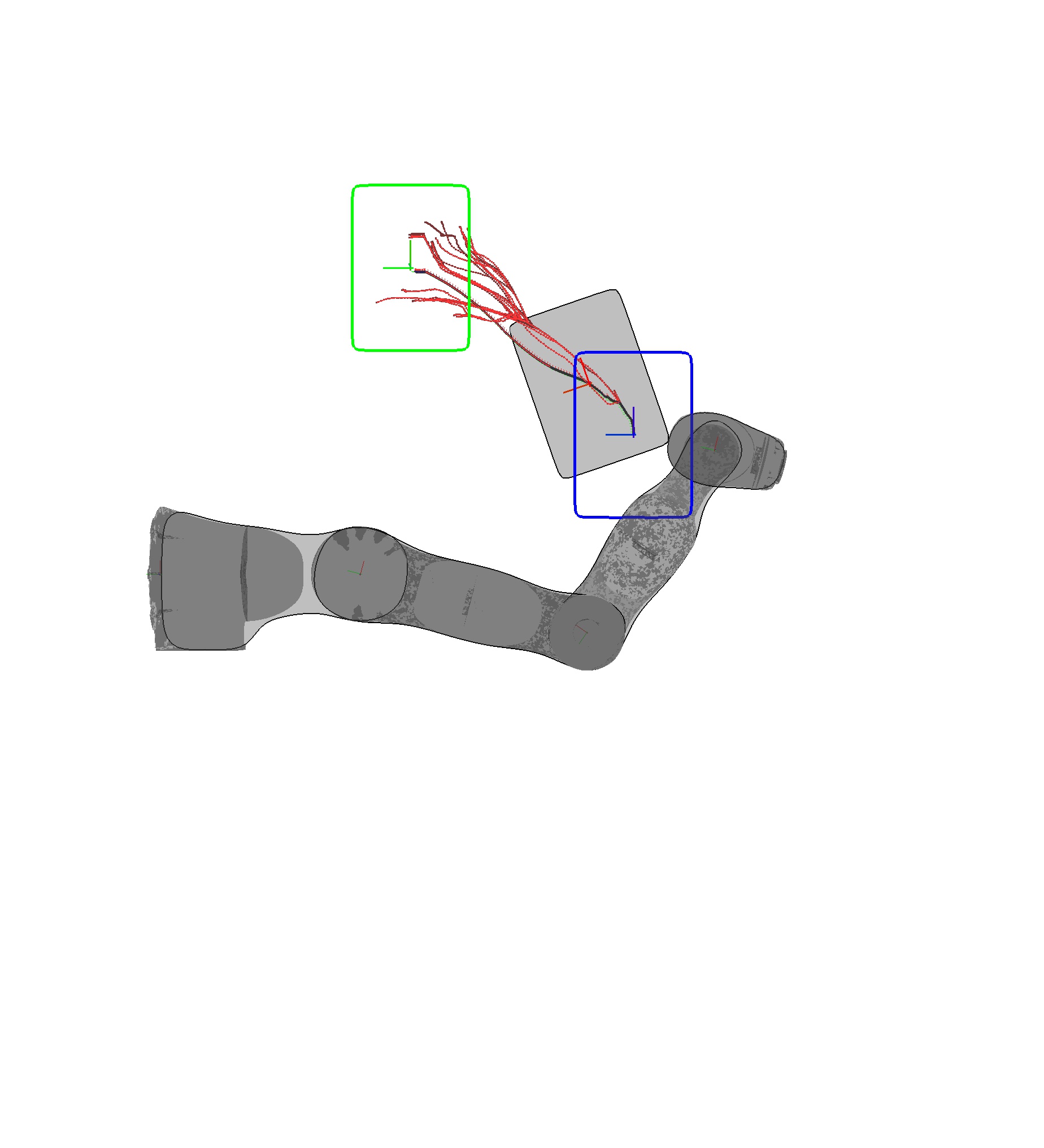}};
            \draw[->, thick] (0.0, 1.2) -- (0.0, 1.5);
        \end{tikzpicture}
    \end{subfigure} 
    \hfill
    \begin{subfigure}[t]{0.32\linewidth} 
        \scriptsize
        \begin{tikzpicture} 
            \node {\includegraphics[width=1\linewidth, trim=255 800 255 225, clip] {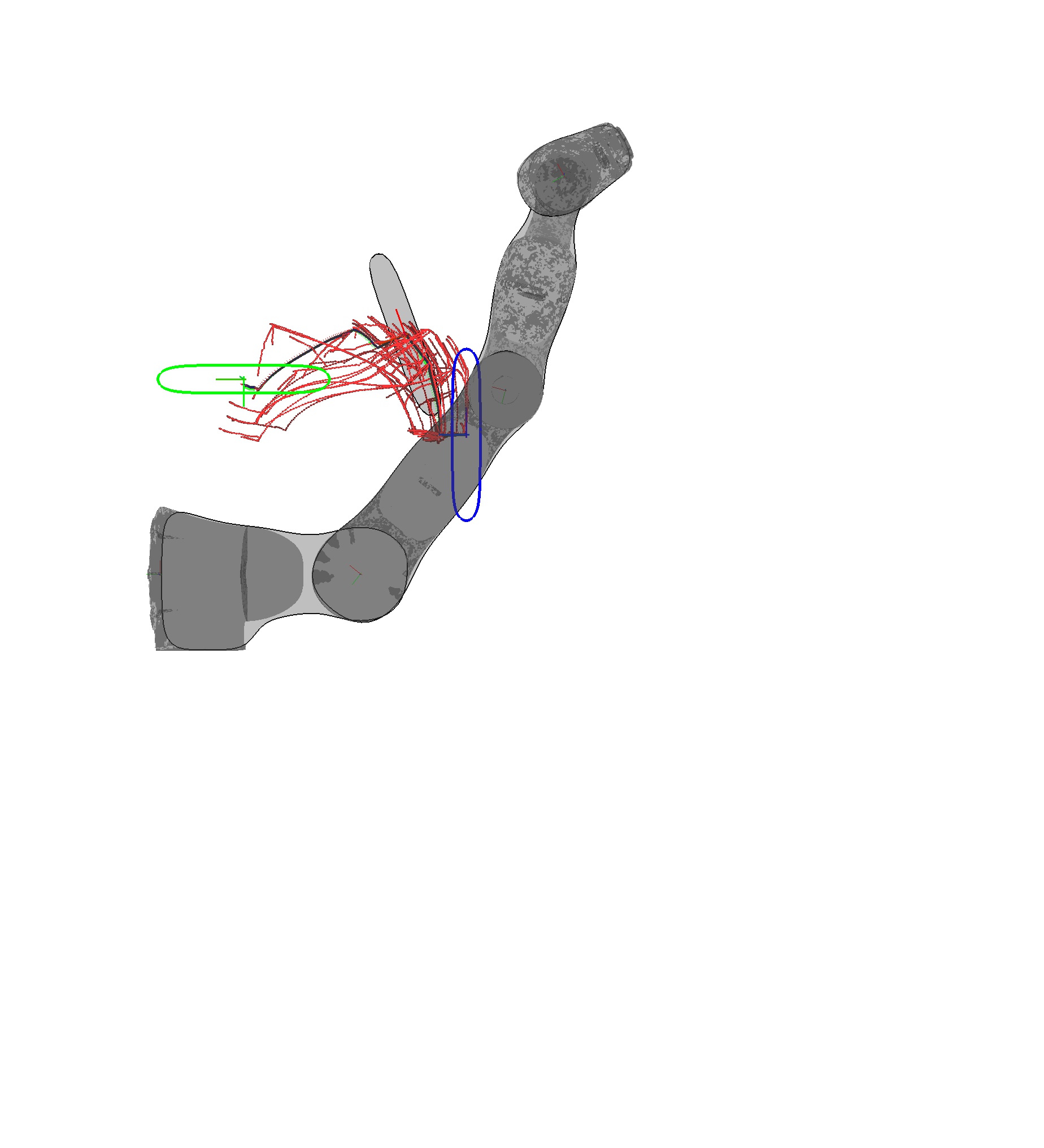}};
            \draw[->, thick] (0.0, 1.2) -- (0.0, 1.5);
        \end{tikzpicture} 
    \end{subfigure} 
    \hfill \vspace{-0.17cm}
    
    \hfill
    \begin{subfigure}[t]{0.32\linewidth} 
        \scriptsize
        \begin{tikzpicture} 
            \node {\includegraphics[width=1\linewidth, trim=255 800 255 255, clip] {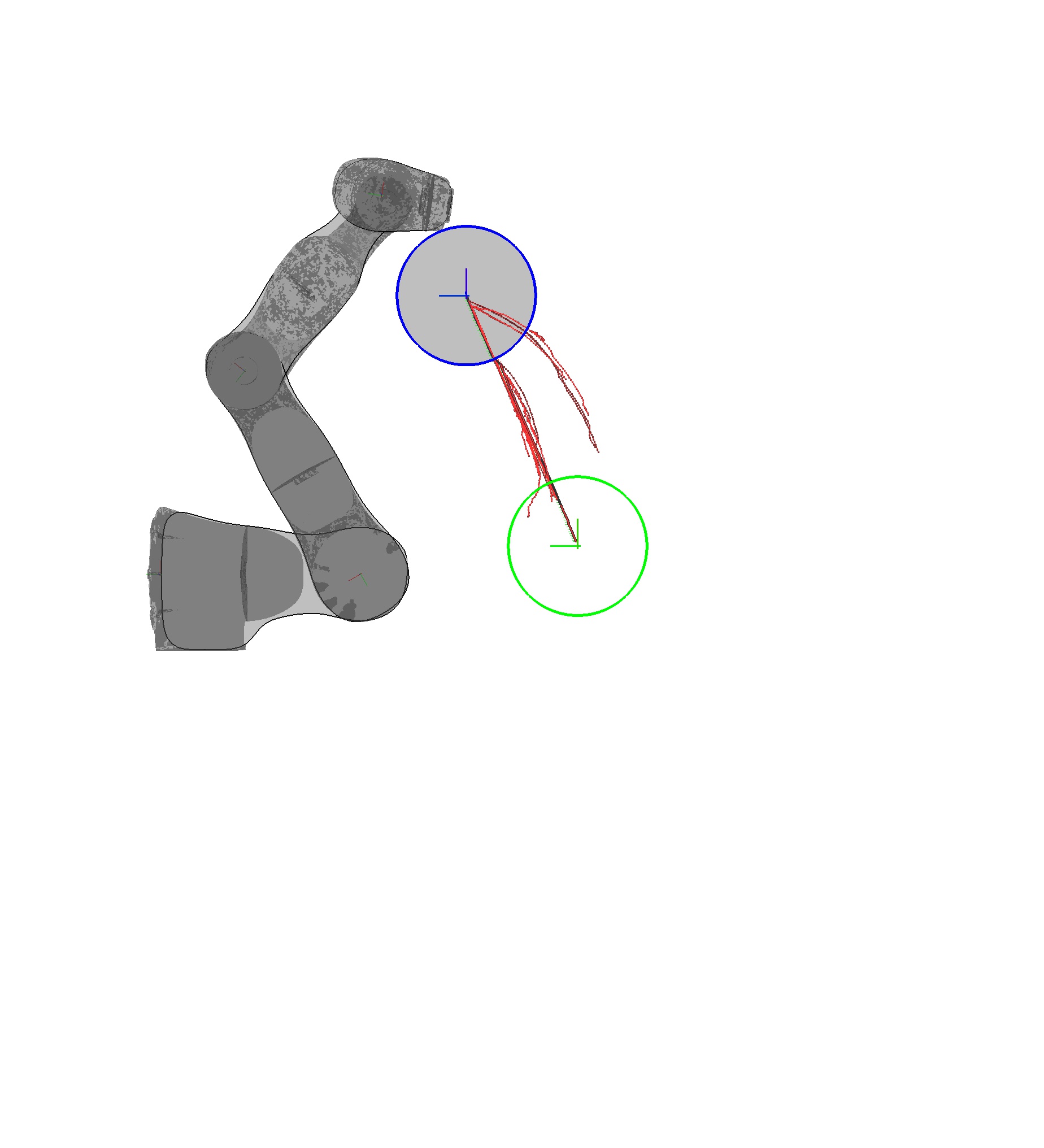}};
            \node at (0.4, 0.6) [color=blue] {$q_u^{init}$};
            \node at (0.7, -0.1) [color=darkgreen] {$q_u^{goal}$};
            \draw[->, thick] (0.0, 1.2) -- (0.0, 1.5);
        \end{tikzpicture}
        \centering
        \captionsetup{justification=centering,margin=0.0cm}
        \caption{Cylinder}
    \end{subfigure} 
    \hfill
    \begin{subfigure}[t]{0.32\linewidth} 
        \scriptsize
        \begin{tikzpicture} 
            \node {\includegraphics[width=1\linewidth, trim=255 800 255 255, clip] {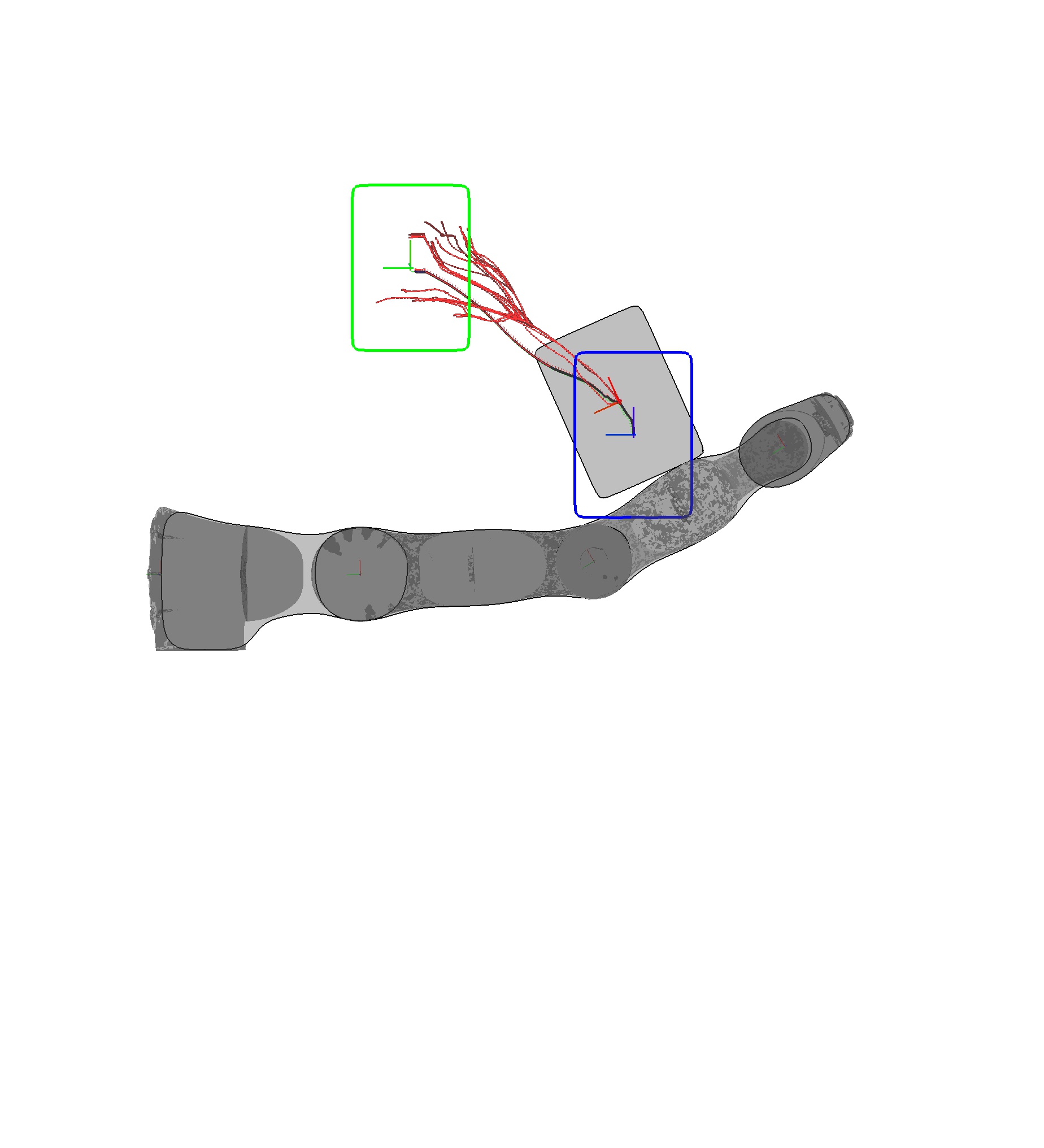}};
            \node at (1.0, 0.3) [color=blue] {$q_u^{init}$};
            \node at (-0.8, 0.0) [color=darkgreen] {$q_u^{goal}$};
            \draw[->, thick] (0.0, 1.2) -- (0.0, 1.5);
        \end{tikzpicture}
        \centering
        \captionsetup{justification=centering,margin=0.0cm}
        \caption{A4 box}
    \end{subfigure} 
    \hfill
    \begin{subfigure}[t]{0.32\linewidth} 
        \scriptsize
        \begin{tikzpicture} 
            \node {\includegraphics[width=1\linewidth, trim=255 800 255 255, clip] {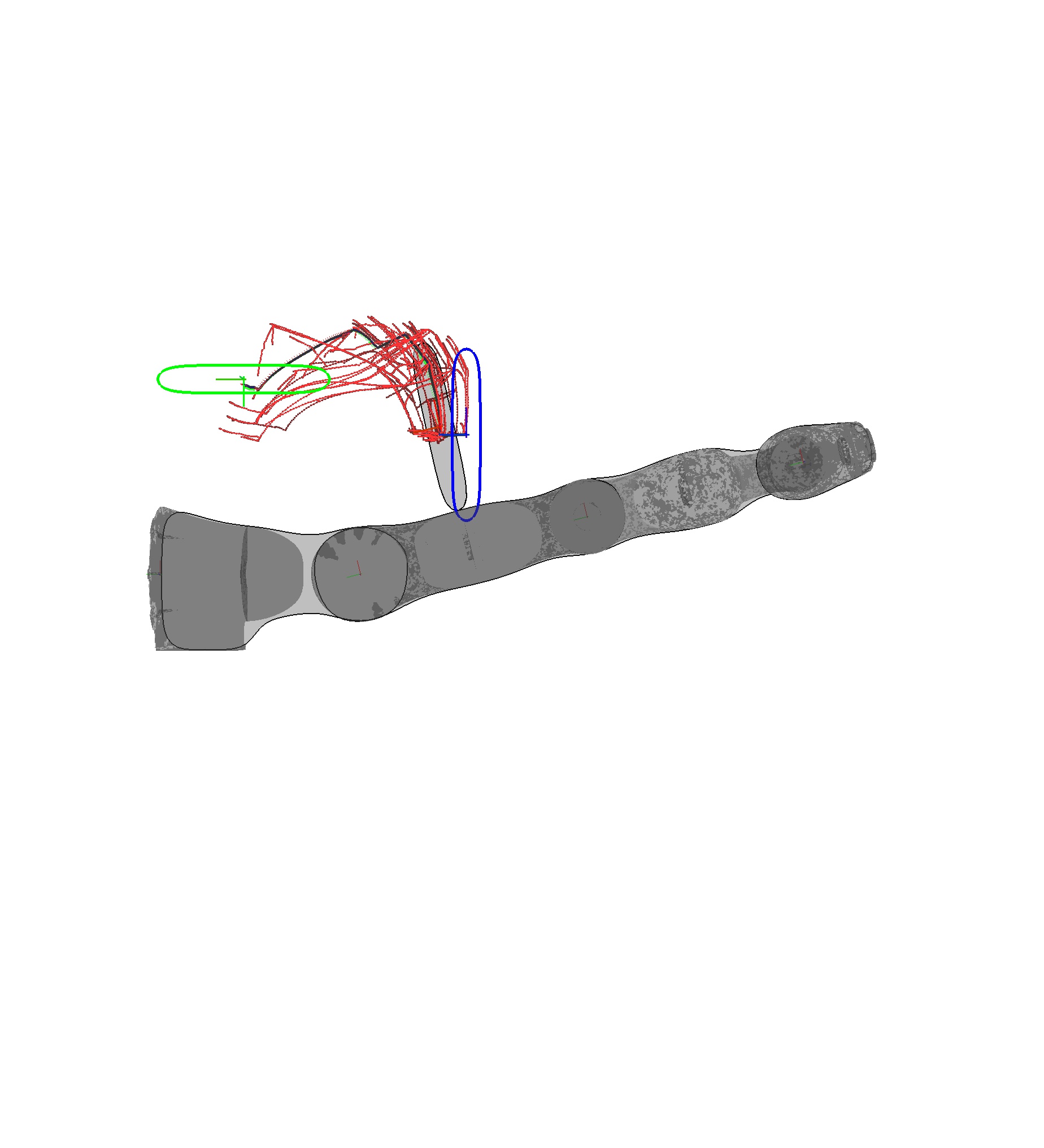}};
            \node at (0.3, 0.3) [color=blue] {$q_u^{init}$};
            \node at (-0.8, 0.6) [color=darkgreen] {$q_u^{goal}$};
            \draw[->, thick] (0.0, 1.2) -- (0.0, 1.5);
        \end{tikzpicture} 
        \centering
        \captionsetup{justification=centering,margin=0.0cm}
        \caption{Capsule}
    \end{subfigure} 
    \hfill
    
    \centering
    \captionsetup{justification=centering,margin=0.0cm}
    \caption{Snapshots and planned trees of various planar Whole-Body Contact-Rich Manipulation (bottom to top)}
    \label{fig:versatility}
\end{figure}

\subsection{Experiment 3: Versatility of manipulation}
\label{sub:exp3}
\subsubsection{Protocol}
To show the versatility of our method in WBCRM scenarios, we conduct pushing planning of different objects from an initial state close to the robot arm and hardly reachable with the last link of the robot, forcing the use of several links to achieve the manipulation. Three types of objects are considered: a cylinder (radius 75mm), an A4 size box ($296\times210$mm) and a capsule ($296\times50$mm). Given the complexity of the planning and to avoid falling into local minima when pushing the object using links with low Degrees of Freedom, for this experiment we input to our \textit{Context Sampling} a randomized subgoal changing at each iteration instead of the fixed final target $q_u^{goal}$.

\subsubsection{Results}
We demonstrate that the planning using different moving links of the robot is possible with our method, and provide some snapshots of these plans in Fig.\ref{fig:versatility}. However, the planning times remain long (40min for the cylinder, 32min for the A4 box and 1h23min for the capsule), mainly because of the difficulty for the planner to find the optimal configuration where to switch the link contacting with the object.
Nevertheless, the presented results prove that our proposed representation is promising for manipulating any object with a convex shape.

\subsection{Transfer to hardware} 

\subsubsection{Protocol}
We conduct an experiment on a real setup consisting in a A4 size box and a KUKA IIWA 14 robot, planning a pushing trajectory to move the box to the target marked on the table in Fig.\ref{fig:exp4_figure} (the green line indicates the orientation) and executing it on the robot.
To cope with the friction uncertainty of the real plant, we allow re-planning a trajectory in a non real-time closed-loop each time the box state differs too much from the planned trajectory. The geometry of the hardware is assumed known, and the box state is tracked using an external motion capture system.

\subsubsection{Results}
First we proved that the robot is able to contact on the box accurately with different links and locations on its non-convex surface, as shown in Fig.\ref{fig:exp4_figure}, acknowledging the suitability of our contact location representation. Second, we succeeded in implementing a non real-time re-planning loop to reach the desired goal despite the uneven friction between the box and the table. This result is shown in the accompanying video. This is an encouraging result towards controlling planar WBCRM in real-time in the future.

\section{CONCLUSIONS}
We have introduced a method relying on a novel explicit representation of the contact surface of the robot and the object to be manipulated. This representation allowed us to frame the problem of contact and motion planning for planar Whole-Body Contact-Rich Manipulation as a set of efficient optimization problems. Our method not only has shown scalability to the added dimensionality of the contact location on the robot outline, but it also yielded a drastic improvement of the planning performance and quality for re-orientation tasks. Moreover, it provides promising foundations for tackling more complex WBCRM tasks using the whole contacting surface of the robot.

Regardless, several limitations and challenges remain. Our method can currently handle one point contact and apply to planar cases where the 2D outline of each link is independent of the joint configuration. We plan to extend our method to 3D cases in the future. Since our pipeline relies extensively on non-convex optimization, it may fall into local minima in some cases. Yet it is possible to sample randomly a subgoal $q_u^{goal}$ at each iteration as in a RRT to help avoiding local minima.
Also, the discrete variables in our problem are chosen randomly and incorporating them into our optimization could improve further the convergence. 
Last but not least, although the feasibility of planar WBCRM has been established and its planning time improved for scenarios with re-orientation, the computation cost associated with non-convexity, non-linearity and high dimensionality of the problem makes it still impossible to implement in a real-time control loop. We see the investigation of possible new representations that enable the implementation of a Model Predictive Control for WBCRM tasks as an interesting future research direction. 
This work also raises the interesting questions of what are the good locations for contact on a robot, what is a good design of the robot links to enhance WBCRM skills, and what are the relevant sensor densities and distributions required to implement a control-loop for WBCRM.




\bibliographystyle{IEEEtran} 
\bibliography{references} 

\end{document}